\documentclass[english]{article}
\usepackage[latin9]{inputenc}
\usepackage{array}
\usepackage{float}
\usepackage{multirow}
\usepackage{amsmath}
\usepackage{amssymb}
\usepackage{graphicx}

\makeatletter

\providecommand{\tabularnewline}{\\}
\floatstyle{ruled}
\newfloat{algorithm}{tbp}{loa}
\providecommand{\algorithmname}{Algorithm}
\floatname{algorithm}{\protect\algorithmname}

\usepackage{iclr2019_conference}

\usepackage{microtype}

\usepackage{xcolor}
\usepackage{algorithmicx,algpseudocode}

\usepackage{hyperref}
\usepackage{url}
\usepackage{fancyhdr, times}

\renewcommand{\cite}{\citep}

\iclrfinalcopy 

\@ifundefined{showcaptionsetup}{}{%
 \PassOptionsToPackage{caption=false}{subfig}}
\usepackage{subfig}
\makeatother

\usepackage{babel}

\global\long\def\sigmoid{\text{sigmoid}}%
\global\long\def\Graph{\mathcal{G}}%
\global\long\def\Vertices{\mathcal{V}}%
\global\long\def\Edges{\mathcal{E}}%
\global\long\def\Neigh{\mathcal{N}}%
\global\long\def\vb{\boldsymbol{v}}%
\global\long\def\eb{\boldsymbol{e}}%
\global\long\def\cb{\boldsymbol{c}}%
\global\long\def\xb{\boldsymbol{x}}%
\global\long\def\hb{\boldsymbol{h}}%
\global\long\def\zb{\boldsymbol{z}}%
\global\long\def\mb{\boldsymbol{m}}%
\global\long\def\rb{\boldsymbol{r}}%
\global\long\def\alphab{\boldsymbol{\alpha}}%
\global\long\def\Real{\mathbb{R}}%
\global\long\def\Loss{\mathcal{L}}%
\global\long\def\Model{\mathrm{GTPN}}%
\global\long\def\StateSet{\mathcal{S}}%
\global\long\def\ActionSet{\mathcal{A}}%

\begin{document}
\title{Graph Transformation Policy Network \\ for Chemical Reaction Prediction }
\author{Kien Do, Truyen Tran and Svetha Venkatesh\\
Applied Artificial Intelligence Institute\\
Deakin University, Geelong, Australia\\
\emph{\{dkdo,truyen.tran,svetha.venkatesh\}@deakin.edu.au}}

\maketitle

\begin{abstract}
We address a fundamental problem in chemistry known as chemical reaction
product prediction. Our main insight is that the input reactant and
reagent molecules can be jointly represented as a graph, and the process
of generating product molecules from reactant molecules can be formulated
as a sequence of graph transformations. To this end, we propose Graph
Transformation Policy Network (GTPN) -- a novel generic method
that combines the strengths of graph neural networks and reinforcement
learning to learn the reactions directly from data with minimal
chemical knowledge. Compared to previous methods, GTPN has some
appealing properties such as: end-to-end learning, and making no assumption
about the length or the order of graph transformations. In order to
guide model search through the complex discrete space of sets of bond
changes effectively, we extend the standard policy gradient loss by
adding useful constraints. Evaluation results show that GTPN
improves the top-1 accuracy over the current state-of-the-art method
by about 3
and prediction errors are also analyzed carefully in the paper.

\end{abstract}

\section{Introduction\label{sec:Introduction}}

Chemical reaction product prediction is a fundamental problem in organic
chemistry. It paves the way for planning syntheses of new substances
\cite{chen2009no}. For decades, huge effort has been spent to solve
this problem. However, most methods still depend on the handcrafted
reaction rules \cite{chen2009no,kayala2011machine,wei2016neural}
or heuristically extracted reaction templates \cite{segler2017neural,coley2017prediction},
thus are not well generalizable to unseen reactions.

\begin{figure}
\begin{centering}
\includegraphics[width=0.95\textwidth]{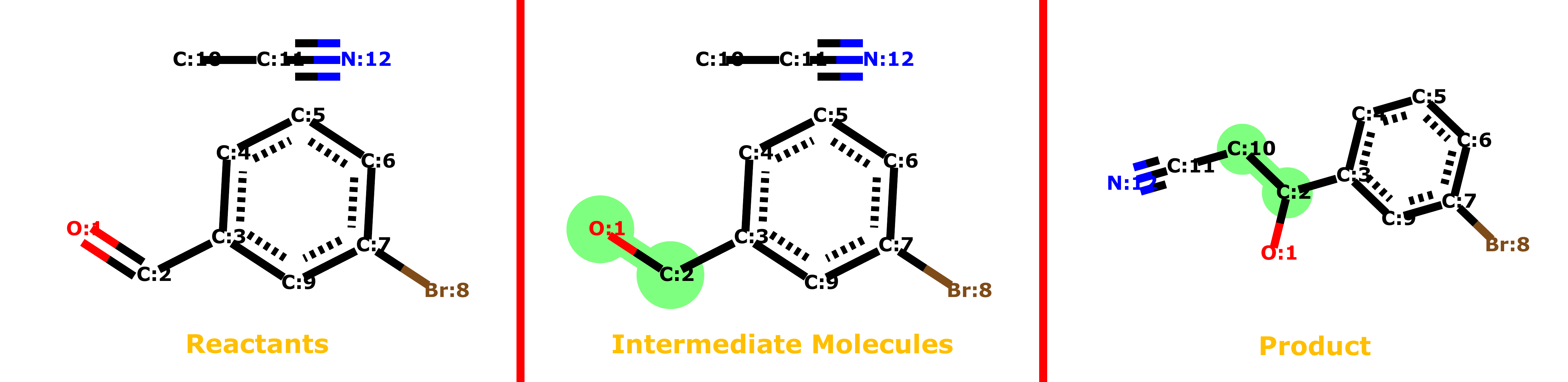}
\par\end{centering}
\caption{A sample reaction represented as a set of graph transformations from
reactants (leftmost) to products (rightmost). Atoms are labeled with
their type (Carbon, Oxygen,...) and their index (1, 2,...) in the
molecular graph. The atom pairs that change connectivity and their
new bonds (if existed) are highlighted in green. There are two bond
changes in this case: 1) The double bond between O:1 and C:2 becomes
single. 2) A new single bond between C:2 and C:10 is added.\label{fig:A-sample-reaction}}
\end{figure}

A reaction can be regarded as a set (or unordered sequence) of graph
transformations in which reactants represented as molecular graphs
are transformed into products by modifying the bonds between some
atom pairs \cite{jochum1980principle,ugi1979new}. See Fig.~\ref{fig:A-sample-reaction}
for an illustration. We call an atom pair $(u,v)$ that changes its
connectivity during reaction and its new bond $b$ a \emph{reaction
triple} $(u,v,b)$. The reaction product prediction problem now becomes
predicting a set of reaction triples given the input reactants and
reagents. We argue that in order to solve this problem well, an intelligent
system should have two key capabilities: (a) \emph{Understanding the
molecular graph structure} of the input reactants and reagents so
that it can identify possible reactivity patterns (i.e., atom pairs
with changing connectivity). (b) \emph{Knowing how to choose from
these reactivity patterns} a correct set of reaction triples to generate
the desired products.

Recent state-of-the-art methods \cite{jin2017predicting,bradshaw2018predicting}
have built the first capability by leveraging graph neural networks
\cite{duvenaud2015convolutional,hamilton2017inductive,pham2017column,gilmer2017neural}.
However, these methods are either unaware of the valid sets of reaction
triples \cite{jin2017predicting} or limited to sequences of reaction
triples with a predefined orders \cite{bradshaw2018predicting}.
The main challenge is that the space of all possible configurations
of reaction triples is extremely large and non-differentiable. Moreover,
a small change in the predicted set of reaction triples can lead to
very different reaction products and a little mistake can produce
invalid prediction.

In this paper, we propose a novel method called Graph Transformation
Policy Network ($\Model$) that addresses the aforementioned challenges.
Our model consists of three main components: a graph neural network
(GNN), a node pair prediction network (NPPN) and a policy network
(PN). Starting from the initial graph of reactant and reagent molecules,
our model iteratively alternates between modeling an input graph using
GNN and predicting a reaction triple using NPPN and PN to generate
a new intermediate graph as input for the next step until it decides
to stop. The final generated graph is considered as the predicted
products of the reaction. Importantly, $\Model$ does not assume any
fixed number or any order of bond changes but learn these properties
itself. One can view $\Model$ as a reinforcement learning (RL) agent
that operates on a complex and non-differentiable space of sets of
reaction triples. To guide our model towards learning a diverse yet
robust-to-small-changes policy, we customize our loss function by
adding some useful constraints to the standard policy gradient loss
\cite{mnih2016asynchronous}.

To the best of our knowledge, $\Model$ is the most generic approach
for the reaction product prediction problem so far in the sense that:
i) It combines graph neural networks and reinforcement learning into
a unified framework and trains everything end-to-end; ii) It does
not use any handcrafted or heuristically extracted reaction rules/templates
to predict the products. Instead, it automatically learns various
types of reactions  from the training data and can generalize to
unseen reactions; iii) It can interpret how the products are formed
via the sequence of reaction triples it generates.

We evaluate $\Model$ on two large public datasets named \emph{USPTO-15k}
and \emph{USPTO}. Our method significantly outperforms all baselines
in the top-1 accuracy, achieving new state-of-the-art results of 82.39\%
and 83.20\% on \emph{USPTO-15k} and \emph{USPTO}, respectively. In
addition, we also provide comprehensive analyses about the performance
of $\Model$ and about different types of errors our model could make.

\section{Method\label{sec:Method}}

\subsection{Chemical Reaction as Markov Decision Process of Graph Transformations}

A reaction occurs when reactant molecules interact with each other
in the presence (or absence) of reagent molecules to form new product
molecules by breaking or adding some of their bonds. Our main insight
is that reaction product prediction can be formulated as predicting
a sequence of such bond changes given the reactant and reagent molecules
as input. A bond change is characterized by the atom pair (\emph{where}
the change happens) and the new bond type (\emph{what} is the change).
We call this atom pair a \emph{reaction atom pair} and call this atom
pair with the new bond type a \emph{reaction triple}. 

More formally, we represent the entire system of input reactant and
reagent molecules as a labeled graph $\Graph=\left(\Vertices,\Edges\right)$
with multiple connected components, each of which corresponds to a
molecule. Nodes in $\Vertices$ are atoms labeled with their atomic
numbers and edges in $\Edges$ are bonds labeled with their bond types.
Given $\Graph$ as input, we predict a sequence of reaction triples
that transforms $\Graph$ into a graph of product molecules $\Graph'$.

As reactions vary in number of transformation steps, we represent
the sequence of reaction triples as $(\xi,u,v,b)^{0},(\xi,u,v,b)^{1},...,(\xi,u,v,b)^{T-1}$
or $(\xi,u,v,b)^{0:T}$ for short. Here $T$ is the maximum number
of steps, $(u,v)$ is a pair of nodes, $b$ is the new edge type of
$(u,v)$, and $\xi$ is a binary signal that indicates the end of
the sequence. If the sequence ends at $T_{\text{end}}<T$, $\xi^{0},...\xi^{T_{\text{end}}-1}$
will be $1$ and $\xi^{T_{\text{end}}},...,\xi^{T-1}$ will be $0$.
At every step $\tau$, if $\xi^{\tau}=1$, we apply the predicted
edge change $(u,v,b)^{\tau}$ on the current graph $\Graph^{\tau}$
to create a new intermediate graph $\Graph^{\tau+1}$ as input for
the next step $\tau+1$. This iterative process of graph transformation
can be formulated as a Markov Decision Process (MDP) characterized
by a tuple $(\StateSet,\ActionSet,P,R,\gamma)$, in which $\StateSet$
is a set of states, $\ActionSet$ is a set of actions, $P$ is a state
transition function, $R$ is a reward function, and $\gamma$ is a
discount factor. Since the process is finite and contains no loop,
we set the discount factor $\gamma$ to be $1$. The rest of the MDP
tuple are defined as follows:
\begin{itemize}
\item \textbf{State}: A state $s^{\tau}\in\StateSet$ is an intermediate
graph $\Graph^{\tau}$ generated at step $\tau$ $(0\le\tau<T)$.
When $\tau=0$, we denote $s^{0}=\Graph^{0}=\Graph$.
\item \textbf{Action}: An action $a^{\tau}\in\ActionSet$ performed at step
$\tau$ is the tuple $(\xi,u,v,b)^{\tau}$. The action is composed
of three consecutive sub-actions: $\xi^{\tau}$, $(u,v)^{\tau}$,
and $b^{\tau}$. If $\xi^{\tau}=0$, our model will ignore the next
sub-actions $(u,v)^{\tau}$ and $b^{\tau}$, and all the future actions
$(\xi,u,v,b)^{\tau+1:T}$. Note that setting $\xi^{\tau}$ to be the
first sub-action is useful in case a reaction does not happen, i.e.,
$\xi^{0}=0$
\item \textbf{State Transition}: If $\xi^{\tau}=1$, the current graph $\Graph^{\tau}$
is modified based on the reaction triple $(u,v,b)^{\tau}$ to generate
a new intermediate graph $\Graph^{\tau+1}$. We do not incorporate
chemical rules such as valency check during state transition because
the current bond change may result in invalid intermediate molecules
$\Graph^{\tau}$, but later, other bond changes may compensate it
to create the valid final products $\Graph^{T_{\text{end}}}$.
\item \textbf{Reward}: We use both immediate rewards and delayed rewards
to encourage our model to learn the optimal policy faster. At every
step $\tau$, if the model predicts $\xi^{\tau}$, $(u,v)^{\tau}$
or $b^{\tau}$ correctly, it will receive a positive reward for each
correct sub-action. Otherwise, a negative reward is given. After the
prediction process has terminated, if the generated products are exactly
the same as the groundtruth products, we give the model a positive
reward, otherwise a negative reward. The concrete reward values are
provided in Appendix~\ref{subsec:Model-Configurations}.
\end{itemize}
\begin{figure}
\begin{centering}
\includegraphics[width=0.7\textwidth]{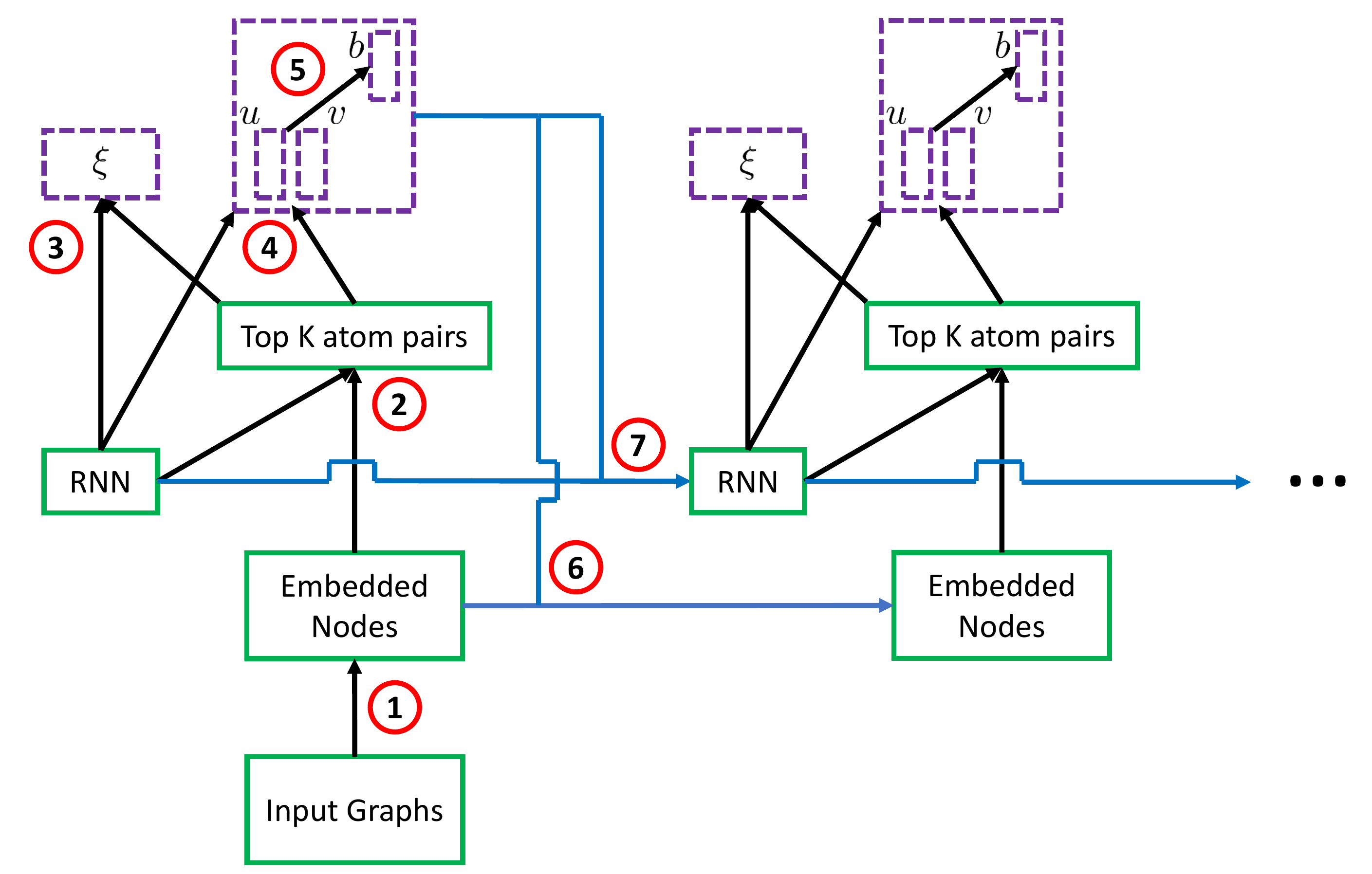}
\par\end{centering}
\caption{Workflow of a Graph Transformation Policy Network ($\protect\Model$).
At every step of the forward pass, our model performs 7 major functions:
1) Computing the atom representation vectors, 2) Computing the most
possible $K$ reaction atom pairs, 3) Predicting the continuation
signal $\xi$, 4) Predicting the reaction atom pair $(u,v)$, 5) Predicting
a new bond $b$ of this atom pair, 6) Updating the atom representation
vectors, and 7) Updating the recurrent state.\label{fig:model_architecture}}
\end{figure}

\subsection{Graph Transformation Policy Network}

In this section, we describe the architecture of our model $-$ a
Graph Transformation Policy Network ($\Model$). $\Model$ has three
main components namely a Graph Neural Network (GNN), a Node Pair Prediciton
Network (NPPN), and a Policy Network (PN). Each component is responsible
for one or several key functions shown in Fig.~\ref{fig:model_architecture}:
GNN performs functions 1 and 6; NPPN performs function 2; and PN performs
functions 3, 4 and 5. Apart from these components, $\Model$ also
has a Recurrent Neural Network (RNN) to keep track of the past transformations.
The hidden state $\hb$ of this RNN is used by NPPN and PN to make
accurate prediction.

\subsubsection{Graph Neural Network}

To model the intermediate graph $\Graph^{\tau}$ at step $\tau$,
we compute the \emph{node state vector} $\xb_{i}^{\tau}$ of every
node $i$ in $\Graph^{\tau}$ by using a variant of the Message Passing
Neural Networks \cite{gilmer2017neural}:

\begin{eqnarray}
\xb_{i}^{\tau} & = & \text{MessagePassing}^{m}\left(\xb_{i}^{\tau-1},\vb_{i},\Neigh^{\tau}(i)\right)\label{eq:atom_repr_vector}
\end{eqnarray}
where $m$ is the number of message passing steps; $\vb_{i}$ is the
feature vector of node $i$; $\Neigh^{\tau}(i)$ is the set of all
neighbor nodes of node $i$; and $\xb_{i}^{\tau-1}$ is the state
vector of node $i$ at the previous step. When $\tau=0$, $\xb_{i}^{\tau-1}$
is initialized from $\vb_{i}$ using a neural network. Details about
the $\text{MessagePassing}(.)$ function are provided in Appendix~\ref{subsec:Message-Passing-Graph}. 

\subsubsection{Node Pair Prediction Network}

In order to predict how likely an atom pair $(i,j)$ of the intermediate
graph $\Graph^{\tau}$ will change its bond, we assign $(i,j)$ with
a score $s_{ij}^{\tau}\in\Real$. If $s_{ij}^{\tau}$ is high, $(i,j)$
is more probably a reaction atom pair, otherwise, less probably. Similar
to \cite{jin2017predicting}, we use two different networks called
\emph{``local''} network and \emph{``global''} network for this
task. In case of the \emph{``local''} network, $s_{ij}^{\tau}$
is computed as:

\begin{eqnarray}
\zb_{ij}^{\tau} & = & \sigma\left(W_{1}\left[\hb^{\tau-1},(\xb_{i}^{\tau}+\xb_{j}^{\tau}),\eb_{ij}\right]+b_{1}\right)\label{eq:np_hid_local}\\
s_{ij}^{\tau} & = & f^{\text{atom pair}}\left(\zb_{ij}^{\tau}\right)\label{eq:np_score_local}
\end{eqnarray}
where $f^{\text{atom pair}}$ is a neural network; $\sigma$ is a
nonlinear activation function (e.g., ReLU); $[.]$ denotes vector
concatenation; $W_{1}$ and $b_{1}$ are parameters; $\hb^{\tau-1}$
is the hidden state of the RNN at the previous step; and $\eb_{ij}$
is the representation vector of the bond between $(i,j)$. If there
is no bond between $(i,j)$ we assume that its bond type is \emph{``NULL''}.
We consider $\zb_{ij}$ as the representation vector for the atom
pair $(i,j)$.

The ``\emph{global}'' network leverages self-attention \cite{vaswani2017attention,wang2018non}
to detect compatibility between atom $i$ and all other atoms before
computing the scores:

\begin{eqnarray}
\rb_{ij}^{\tau} & = & \sigma\left(V_{1}\left[(\xb_{i}^{\tau}+\xb_{j}^{\tau}),\eb_{ij}\right]+c_{1}\right)\nonumber \\
a_{ij}^{\tau} & = & \text{softmax}\left(V_{2}\rb_{ij}^{\tau}+c_{2}\right)\nonumber \\
\cb_{i}^{\tau} & = & \sum_{j\in\Vertices}a_{ij}\xb_{j}^{\tau}\nonumber \\
\zb_{ij}^{\tau} & = & \sigma\left(W_{1}\left[\hb^{\tau-1},(\xb_{i}^{\tau}+\xb_{j}^{\tau}),(\cb_{i}^{\tau}+\cb_{j}^{\tau}),\eb_{ij}\right]+b_{1}\right)\label{eq:np_hid_global}\\
s_{ij}^{\tau} & = & f^{\text{atom pair}}\left(\zb_{ij}^{\tau}\right)\label{eq:np_score_global}
\end{eqnarray}
where $a_{ij}$ is the attention score from node $i$ to every other
node $j$; $\cb_{i}$ is the context vector of atom $i$ that summarizes
the information from all other atoms.

During experiments, we tried both options mentioned above and saw
that the ``global'' network clearly outperforms the \emph{``local''}
network so we set the \emph{``global''} network as a default module
in our model. In addition, since reagents never change their form
during a reaction, we explicitly exclude all atom pairs that have
either atoms belong to the reagents. This leads to better results
than not using reagent information. Detailed analyses are provided
in Appendix~\ref{subsec:Using-Reagents-Information}.

\paragraph{Top-$K$ atom pairs}

Because the number of atom pairs that actually participate in a reaction
is very small (usually smaller than 10) compared to the total number
of atom pairs of the input molecules (usually hundreds or thousands),
it is much more efficient to identify reaction triples from a small
subset of highly probable reaction atom pairs. For that reason, we
extract $K$ $(K\ll|\Vertices|^{2})$ atom pairs with the highest
scores. Later, we will predict reaction triples taken from these $K$
atom pairs only. We denote the set of top-$K$ atom pairs, their corresponding
scores, and representation vectors as $\left\{ (u_{k},v_{k})|k=\overline{1,K}\right\} $,
$\left\{ s_{u_{k}v_{k}}|k=\overline{1,K}\right\} $ and $Z_{K}=\left\{ \zb_{u_{k}v_{k}}|k=\overline{1,K}\right\} $,
respectively.

\subsubsection{Policy Network}

\paragraph{Predicting continuation signal}

To account for varying number of transformation steps, PN generates
a continuation signal $\xi^{\tau}\in\{0,1\}$ to indicate whether
prediction should continue or terminate. $\xi^{\tau}$ is drawn from
a Bernoulli distribution:
\begin{eqnarray}
p\left(\xi^{\tau}=1\right) & = & \sigmoid\left(f^{\text{signal}}\left(\left[\hb^{\tau-1},g\left(Z_{K}^{\tau}\right)\right]\right)\right)\label{eq:signal_prob}
\end{eqnarray}
where $\hb^{\tau-1}$ is the previous RNN state; $Z_{K}^{\tau}$ is
the set of representation vectors of the top $K$ atom pairs at the
current step; $f^{\text{signal}}$ is a neural network; $g$ is a
function that maps an unordered set of inputs to an output vector.
For simplicity, we use a mean function:

\[
\zb_{K}^{\tau-1}=g\left(Z_{K}^{\tau}\right)=\frac{1}{K}\sum_{k=1}^{K}W\zb_{u_{k}v_{k}}^{\tau-1}
\]

\paragraph{Predicting atom pair}

At the next sub-step, PN predicts which atom pair changes its bond
during the reaction by sampling from the top-$K$ atom pairs with
probability:

\begin{equation}
p\left((u_{k},v_{k})^{\tau}\right)=\text{softmax}_{K}\left(s_{u_{k}v_{k}}^{\tau}\right)\label{eq:atom_pair_prob}
\end{equation}
where $s_{u_{k}v_{k}}^{\tau}$ is the score of the atom pair $(u_{k},v_{k})^{\tau}$
computed in Eq.~(\ref{eq:np_score_global}). After predicting the
atom pair $(u,v)^{\tau}$, we will mask it to ensure that it could
not be in the top $K$ again at future steps. 

\paragraph{Predicting bond type}

Given an atom pair $(u,v)^{\tau}$ sampled from the previous sub-step,
we predict a new bond type $b^{\tau}$ between $u$ and $v$ to get
a complete reaction triple $(u,v,b)^{\tau}$ using the probability:

\begin{equation}
p\left(b^{\tau}|(u,v)^{\tau}\right)=\text{softmax}_{B}\left(f^{\text{bond}}\left(\left[\hb^{\tau-1},\zb_{uv}^{\tau},\left(\eb_{b}-\eb_{b^{\text{old}}}\right)\right]\right)\right)\label{eq:bond_prob}
\end{equation}
where $B$ is the total number of bond types; $\zb_{uv}^{\tau}$ is
the representation vector of $(u,v)^{\tau}$ computed in Eq.~(\ref{eq:np_hid_global});
$b^{\text{old}}$ is the old bond of $(u,v)$; $\eb_{b^{\text{old}}}$
and $\eb_{b}$ are the embedding vectors corresponding to the bond
type $b^{\text{old}}$ and $b$, respectively; and $f^{\text{bond}}$
is a neural network.

\subsection{Updating States}

After predicting a complete reaction triple $(u,v,b)^{\tau}$, our
model updates: i) the new recurrent hidden state $\hb^{\tau}$, and
ii) the new node representation vectors $\xb_{i}^{\tau+1}$ of the
new intermediate graph $\Graph^{\tau+1}$ for $i\in\Vertices$. These
updates are presented in Appendix~\ref{subsec:Updating-States}.

\subsection{Training}

Loss function plays a central role in achieving fast training and
high performance. We design the following loss:

\[
\Loss=\lambda_{1}\Loss^{\text{A2C}}+\lambda_{2}\Loss^{\text{value}}+\lambda_{3}\Loss^{\text{atom pair}}+\lambda_{4}\Loss^{\text{over length}}+\lambda_{5}\text{\ensuremath{\Loss}}^{\text{in top }K}
\]
where $\Loss^{\text{A2C}}$ is the Advantage Actor-Critic (A2C) loss
\cite{mnih2016asynchronous} to account for the correct sequence of
reaction triples; $\Loss^{\text{value}}$ is the loss for estimating
the value function used in A2C; $\Loss^{\text{atom pair}}$ accounts
for binary change in the bond of an atom pair; $\Loss^{\text{over length}}$
penalizes long predicted sequences; and $\text{\ensuremath{\Loss}}^{\text{in top }K}$
is the rank loss to force a ground-truth reaction atom pair to appear
in the top-$K$; and $\lambda_{1},...,\lambda_{5}>0$ are tunable
coefficients. The component losses are explained in the following.

\subsubsection{Reaction triple loss}

The loss follows a policy gradient method known as Advantage Actor-Critic
(A2C):

\begin{eqnarray}
\Loss^{\text{A2C}} & = & -\sum_{\tau=0}^{T_{\text{end}}-1}\left(A_{\text{signal}}^{\tau}\log p\left(\xi^{\tau}\right)+A_{\text{atom pair}}^{\tau}\log p\left((u,v)^{\tau}\right)+A_{\text{bond}}^{\tau}\log p\left(b^{\tau}\right)\right)\nonumber \\
 &  & -A_{\text{signal}}^{T_{\text{end}}}\log\pi\left(\xi^{T_{\text{end}}}\right)\label{eq:actor_critic_loss}
\end{eqnarray}
where $T_{\text{end}}$ is the first step that $\xi=0$; $A_{\text{signal}}$,
$A_{\text{atom pair}}$ and $A_{\text{bond}}$ are called \emph{advantages}.
To compute these advantages, we use the unbiased estimations called
Temporal Different errors, defined as:

\begin{eqnarray}
A_{\text{signal}}^{\tau} & = & r_{\text{signal}}^{\tau}+\gamma V_{\phi}\left(Z_{K}^{\tau+1}\right)-V_{\phi}\left(Z_{K}^{\tau}\right)\label{eq:TD_error_signal}\\
A_{\text{atom pair}}^{\tau} & = & r_{\text{atom pair}}^{\tau}+\gamma V_{\phi}\left(Z_{K}^{\tau+1}\right)-V_{\phi}\left(Z_{K}^{\tau}\right)\label{eq:TD_error_uv}\\
A_{\text{bond}}^{\tau} & = & r_{\text{bond}}^{\tau}+\gamma V_{\phi}\left(Z_{K}^{\tau+1}\right)-V_{\phi}\left(Z_{K}^{\tau}\right)\label{eq:TD_error_b}
\end{eqnarray}
where $r_{\text{signal}}^{\tau}$, $r_{\text{atom pair}}^{\tau}$,
$r_{\text{bond}}^{\tau}$ are immediate rewards at step $\tau$; at
the final step $\tau=T_{\text{end}}$, the model receives additional
delayed rewards; $\gamma$ is the discount factor; and $V_{\phi}$
is the parametric value function. We train $V_{\phi}$ using the following
mean square error loss:

\begin{eqnarray}
\Loss^{\text{value}} & = & \sum_{\tau=0}^{T_{\text{end}}}\left\Vert V_{\phi}\left(Z_{K}^{\tau}\right)-R^{\tau}\right\Vert ^{2}\label{eq:value_loss}
\end{eqnarray}
where $R^{\ensuremath{\tau}}$ is the return at step $\tau$.

\paragraph{Episode termination during training}

Although the loss defined in Eq.~(\ref{eq:actor_critic_loss}) is
correct, it is not good to use in practice because: i) If our model
selects a wrong sub-action at any sub-step of the step $T_{\text{wrong}}$
($T_{\text{wrong}}<T_{\text{end}}$), the whole predicted sequence
will be incorrect regardless of what will be predicted from $T_{\text{wrong}}+1$
to $T_{\text{end}}$. Therefore, computing the loss for actions from
$T_{\text{wrong}}+1$ to $T_{\text{end}}$ is redundant. ii) More
importantly, the incorrect updates of the graph structure at subsequent
steps from $T_{\text{wrong}}+1$ to $T_{\text{end}}$ will lead to
cumulative prediction errors which make the training of our model
much more difficult.

To resolve this issue, during training, we use a binary vector $\boldsymbol{\zeta}\in\{0,1\}^{3T}$
to keep track of the first wrong sub-action: $\zeta^{t}=\begin{cases}
1 & \text{if}\ t\leq t_{\text{first wrong}}\\
0 & \text{if}\ t>t_{\text{first wrong}}
\end{cases}$ where $t_{\text{first wrong}}$ denotes the sub-step at which our
model chooses a wrong sub-action the first time. The actor-critic
loss in Eq.~(\ref{eq:actor_critic_loss}) now becomes:

\begin{equation}
\Loss^{\text{A2C}}=-\sum_{\tau=0}^{T}\left(\zeta^{\tau}A_{\text{signal}}^{\tau}\log p\left(\xi^{\tau}\right)+\zeta^{(\tau+1)}A_{\text{atom pair}}^{\tau}\log p\left((u,v)^{\tau}\right)+\zeta^{(\tau+2)}A_{\text{bond}}^{\tau}\log p\left(b^{\tau}\right)\right)\label{eq:actor_critic_loss3}
\end{equation}
where $T$ is the maximum number of steps. Similarly, we change the
value loss into:

\[
\Loss^{\text{value}}=\sum_{\tau=0}^{T}\zeta^{\tau}\left\Vert V_{\phi}\left(Z_{K}^{\tau}\right)-R^{\tau}\right\Vert ^{2}
\]

\subsubsection{Reaction atom pair loss}

To train our model to assign higher scores to reaction atom pairs
and lower to non-reaction atom pairs, we use the following cross-entropy
loss function:

\begin{equation}
\Loss^{\text{atom pair}}=-\sum_{\tau=0}^{T_{\text{first wrong}}}\sum_{i\in\Vertices}\sum_{j\in\Vertices,j\neq i}\eta_{ij\tau}\left(y_{ij}\log p_{ij}+(1-y_{ij})\log(1-p_{ij})\right)\label{eq:atom_pair_loss}
\end{equation}
where $T_{\text{first wrong}}=\left\lfloor \frac{t_{\text{first wrong}}}{3}\right\rfloor $;
$\eta_{ijt}\in\{0,1\}$ is a mask of the atom pair $(i,j)$ at step
$\tau$; $y_{ij}\in\{0,1\}$ is the label indicating whether the atom
pair $(i,j)$ is a reaction atom pair or not; $p_{\ensuremath{ij}}=\sigmoid(s_{ij})$
(see Eq.~(\ref{eq:np_score_global})).

\subsubsection{Constraint on the sequence length}

One major difficulty of the chemical reaction prediction problem is
to know exactly when to stop prediction so we can make accurate inference.
By forcing the model to stop immediately when making wrong prediction,
we can prevent cumulative error and significantly reduce variance
during training. But it also comes with a cost: The model cannot learn
(because it does not have to learn) when to stop. This phenomenon
can be visualized easily as the model predicts $1$ for the signal
at every step $\tau$ during inference. In order to make the model
aware of the correct sequence length during training, we define a
loss that punishes the model if it produces a longer sequence than
the ground truth sequence:

\begin{equation}
\Loss^{\text{over length}}=-\sum_{T_{\text{end}}^{\text{gt}}\le\tau<T_{\text{end}}}\log p\left(\xi^{\tau}=0\right)\label{eq:length_constraint_loss}
\end{equation}
where $T_{\text{end}}^{\text{gt}}$ is the end step of the ground-truth
sequence. Note that the loss in Eq.~(\ref{eq:length_constraint_loss})
is not applied when $T_{\text{end}}\le T_{\text{end}}^{\text{gt}}$.
The reason is that forcing $\xi^{\tau}=1$ with $T_{\text{end}}\le\tau<T_{\text{end}}^{\text{gt}}$
is not theoretically correct because all the signals after $T_{\text{end}}$
are assumed to be $0$. The incentive to force $T_{\text{end}}$ close
to $T_{\text{end}}^{\text{gt}}$ when it is smaller than $T_{\text{end}}^{\text{gt}}$
has already been included in the advantages in Eq.~(\ref{eq:actor_critic_loss3}).

\subsubsection{Constraint on the top-$K$ atom pairs}

Ideally, the loss from Eq.~(\ref{eq:atom_pair_loss}) pushes a reaction
atom pair $(\tilde{u},\tilde{v})^{\tau}$ into the top-$K$ atom pairs
at each step $\tau<T_{\text{end}}^{\text{gt}}$. However, this is
not guaranteed, especially when $\tau$ comes close to $T_{\text{end}}^{\text{gt}}$.
To encourage the ground-truth reaction atom pair $(\tilde{u},\tilde{v})^{\tau}$
with the highest score to appear in the top $K$, we introduce an
additional rank-based loss:

\[
\Loss^{\text{in top }K}=-\sum_{\tau=0}^{T_{\text{first wrong}}}\log p\left((\tilde{u},\tilde{v})^{\tau}\ \text{in top }K\right)
\]
where $p\left((\tilde{u},\tilde{v})^{\tau}\ \text{in top }K\right)$
is computed as:

\begin{eqnarray}
p\left((\tilde{u},\tilde{v})^{\tau}\ \text{in top }K\right) & = & \frac{\exp\left(s_{\tilde{u}\tilde{v}}^{\tau}\right)}{\exp\left(s_{\tilde{u}\tilde{v}}^{\tau}\right)+\sum_{k=1}^{K}\exp\left(s_{u_{k}v_{k}}^{\tau}\right)}\label{eq:in_topK_prob}
\end{eqnarray}

\section{Experiments\label{sec:Experiments}}

\subsection{Dataset}

We evaluate our model on two standard datasets \emph{USPTO-15k} (15K
reactions) and \emph{USPTO} (480K reactions) which have been used
in previous works \cite{jin2017predicting,schwaller2017found,bradshaw2018predicting}.
Details about these datasets are given in Table~\ref{tab:Datasets}.
The USPTO dataset contains reactant, reagent and product molecules
represented as SMILES strings. Using RDKit\footnote{https://www.rdkit.org/},
we convert the SMILES strings into molecule objects and store them
as graphs. For each reaction, every atom in the reactant and reagent
molecules is identified with a unique ``atom map number''. This
identity is the same in the products. Using this knowledge, we compare
every atom pair in the input molecules with the correspondent in the
product molecules to obtain a ground-truth set of reaction triples
for training. In \emph{USPTO-15k}, the ground-truth sets of reaction
triples was precomputed by \cite{jin2017predicting}. 

\begin{table}
\begin{centering}
\begin{tabular}{|c|c|c|c|c|c|c|}
\hline 
\multicolumn{2}{|c|}{Dataset} & \emph{\#reactions} & \emph{\#changes} & \emph{\#molecules} & \emph{\#atoms} & \emph{\#bonds}\tabularnewline
\hline 
\hline 
\multirow{3}{*}{\emph{USPTO-15k}} & train & 10,500 & 1 \textbar{} 11 \textbar{} 2.3 & 1 \textbar{} 20 \textbar{} 3.6 & 4 \textbar{} 100 \textbar{} 34.9 & 3 \textbar{} 110 \textbar{} 34.7\tabularnewline
\cline{2-7} 
 & valid & 1,500 & 1 \textbar{} 11 \textbar{} 2.3 & 1 \textbar{} 20 \textbar{} 3.6 & 7 \textbar{} 94 \textbar{} 34.5 & 5 \textbar{} 99 \textbar{} 34.2\tabularnewline
\cline{2-7} 
 & test & 3,000 & 1 \textbar{} 11 \textbar{} 2.3 & 1 \textbar{} 16 \textbar{} 3.6 & 7 \textbar{} 98 \textbar{} 34.9 & 5 \textbar{} 102 \textbar{} 34.7\tabularnewline
\hline 
\hline 
\multirow{3}{*}{\emph{USPTO}} & train & 409,035 & 1 \textbar{} 6 \textbar{} 2.2 & 2 \textbar{} 29 \textbar{} 4.8 & 9 \textbar{} 150 \textbar{} 39.7 & 6 \textbar{} 165 \textbar{} 38.6\tabularnewline
\cline{2-7} 
 & valid & 30,000 & 1 \textbar{} 6 \textbar{} 2.2 & 2 \textbar{} 25 \textbar{} 4.8 & 9 \textbar{} 150 \textbar{} 39.6 & 7 \textbar{} 158 \textbar{} 38.5\tabularnewline
\cline{2-7} 
 & test & 40,000 & 1 \textbar{} 6 \textbar{} 2.2 & 2 \textbar{} 22 \textbar{} 4.8 & 9 \textbar{} 150 \textbar{} 39.8 & 7 \textbar{} 162 \textbar{} 38.7\tabularnewline
\hline 
\end{tabular}
\par\end{centering}
\caption{Statistics of \emph{USPTO-15k} and \emph{USPTO} datasets. \emph{``changes''}
means bond changes, \emph{``molecules''} means reactants and reagents
in a reaction; \emph{``atoms''} and \emph{``bonds''} are defined
for a molecule. Apart from \emph{``\#reactions''}, other columns
are presented in the format ``min \textbar{} max \textbar{} mean''.\label{tab:Datasets}}
\end{table}

\subsection{Reaction Atom Pair Prediction\label{subsec:Reaction-Atom-Pair}}

In this section, we test our model's ability to identify reaction
atom pairs by formulating it as a ranking problem with the scores
computed in Eq.~(\ref{eq:np_score_global}). Similar to \cite{jin2017predicting},
we use \emph{Coverage@k} as the evaluation metric, which is the proportion
of reactions that have \emph{all} groundtruth reaction atom pairs
appear in the top $k$ predicted atom pairs.

We compare our proposed graph neural network (GNN) with Weisfeiler-Lehman
Network (WLN) \cite{jin2017predicting} and Column Network (CLN) \cite{pham2017column}.
Since our GNN explicitly uses reagent information to compute the scores
of atom pairs, we modify the implementation of WLN and CLN accordingly
for fair comparison. From Table~\ref{tab:atom_pair_prediction},
we observe that our GNN clearly outperforms WLN and CLN in all cases.
We attribute this improvement to the use of a separate node state
vector $\xb_{i}^{t}$ (different from the node feature vector $\vb_{i}$)
for updating the structural information of a node (see Eq.~(\ref{eq:generic_node_update})).
The other two models, on the other hand, only use a single vector
to store both the node features and structure, hence, some information
may be lost. In addition, using explicit reagent information boosts
the prediction accuracy, which improves the WLN by 1-7\% depending
on the metrics. The presence of reagent information reduces the number
of atom pairs to be searched on and contributes to the likelihood
of reaction atom pairs. Further results are presented in Appendix~\ref{subsec:Using-Reagents-Information}.

\begin{table}
\begin{centering}
\begin{tabular}{|c|c|c|c||c|c|c|}
\hline 
\multirow{2}{*}{Model} & \multicolumn{3}{c||}{\emph{USPTO-15k}} & \multicolumn{3}{c|}{\emph{USPTO}}\tabularnewline
\cline{2-7} 
 & \emph{C@6} & \emph{C@8} & \emph{C@10} & \emph{C@6} & \emph{C@8} & \emph{C@10}\tabularnewline
\hline 
\hline 
WLN$^{\star}$ \cite{jin2017predicting} & 81.6 & 86.1 & 89.1 & 89.8 & 92.0 & 93.3\tabularnewline
\hline 
WLN \cite{jin2017predicting} & 88.45 & 91.65 & 93.34 & 90.97 & 93.98 & 95.26\tabularnewline
\hline 
CLN \cite{pham2017column} & 88.68 & 91.63 & 93.07 & 90.72 & 93.57 & 94.80\tabularnewline
\hline 
Our GNN & \textbf{88.92} & \textbf{92.00} & \textbf{93.57} & \textbf{91.24} & \textbf{94.17} & \textbf{95.33}\tabularnewline
\hline 
\end{tabular}
\par\end{centering}
\caption{Results for reaction atom pair prediction. \emph{C@k} is coverage
at $k$. Best results are highlighted in bold. WLN$^{\star}$ is the
original model from \cite{jin2017predicting} while WLN is our re-implemented
version. Except for WLN$^{\star}$, other models explicitly use reagent
information.\label{tab:atom_pair_prediction}}
\end{table}

\subsection{Top-$K$ Atom Pair Extraction}

\begin{figure}
\begin{centering}
\includegraphics[width=0.5\textwidth]{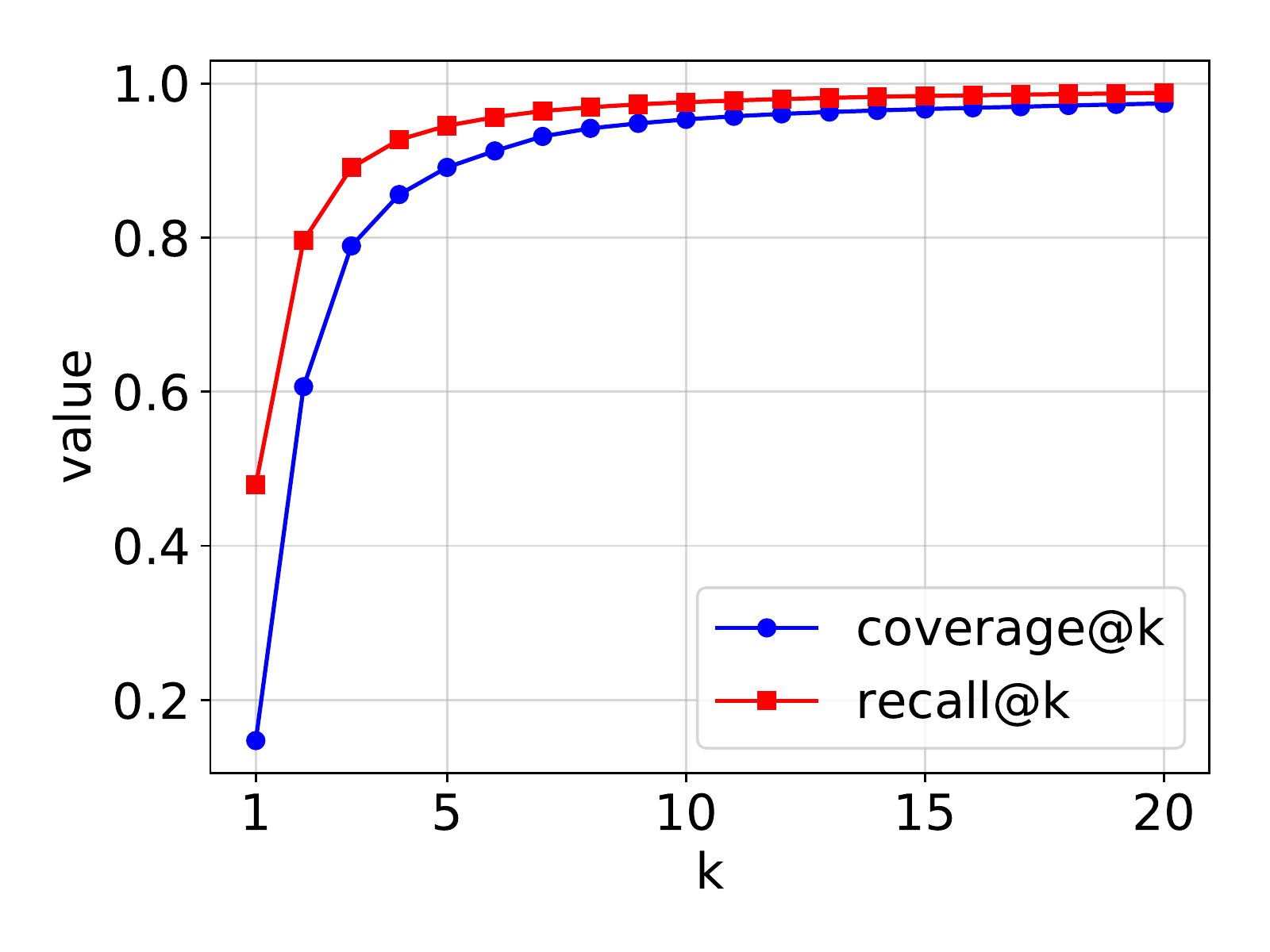}
\par\end{centering}
\caption{\emph{Coverage@k} and \emph{Recall@k} with respect to $k$ for the
\emph{USPTO} dataset.\label{fig:Coverage_Recall}}
\end{figure}
The performance of our model depends on the number of selected top
atom pairs $K$. The value of $K$ presents a trade-off between coverage
and efficiency. In addition to the metric \emph{Coverage@k} in Sec.~\ref{subsec:Reaction-Atom-Pair},
we use \emph{Recall@k} which is the proportion of correct atom pairs
that appear in top \emph{$k$} to find the good $K$. Fig.~\ref{fig:Coverage_Recall}
shows \emph{Coverage@k} and \emph{Recall@k }for the \emph{USPTO} dataset
with respect to $k$. We see that both curves increase rapidly when
$k<10$ and stablize when $k>10$. We also ran experiments with $k=10$,
$15$, $20$ and observed that their prediction results are quite
similar. Hence, in what follows we select $K=10$ for efficiency.

\subsection{Reaction Product Prediction\label{subsec:Reaction-Product-Prediction}}

\begin{table}
\begin{centering}
\begin{tabular}{|c|c|c|c||c|c|c|}
\hline 
\multirow{2}{*}{Model} & \multicolumn{3}{c||}{\emph{USPTO-15k}} & \multicolumn{3}{c|}{\emph{USPTO}}\tabularnewline
\cline{2-7} 
 & \emph{P@1} & \emph{P@3} & \emph{P@5} & \emph{P@1} & \emph{P@3} & \emph{P@5}\tabularnewline
\hline 
\hline 
WLDN \cite{jin2017predicting} & \emph{76.7} & \emph{85.6} & \textbf{\emph{86.8}} & \emph{79.6} & \textbf{\emph{87.7}} & \textbf{\emph{89.2}}\tabularnewline
\hline 
Seq2Seq \cite{schwaller2017found} & - & - & - & 80.3$^{\star}$ & 86.2$^{\star}$ & 87.5$^{\star}$\tabularnewline
\hline 
\hline 
$\Model$ & 72.31 & - & - & 71.26 & - & -\tabularnewline
\hline 
$\Model$$^{\diamondsuit}$ & 74.56 & 82.62 & 84.23 & 73.25 & 80.56 & 83.53\tabularnewline
\hline 
$\Model$$^{\diamondsuit\clubsuit}$ & 74.56 & 83.19 & 84.97 & 73.25 & 84.31 & 85.76\tabularnewline
\hline 
$\Model$$^{\diamondsuit\spadesuit}$ & \textbf{82.39} & 85.60 & 86.68 & \textbf{83.20} & 84.97 & 85.90\tabularnewline
\hline 
$\Model$$^{\diamondsuit\spadesuit\clubsuit}$ & \textbf{82.39} & \textbf{85.73} & \textbf{86.78} & \textbf{83.20} & 86.03 & 86.48\tabularnewline
\hline 
\end{tabular}
\par\end{centering}
\caption{Results for reaction prediction. \emph{P@k} is precision at $k$.
State-of-the-art results from \cite{jin2017predicting} are written
in italic. Results from \cite{schwaller2017found} are marked with
$^{\star}$ and they are computed on a slightly different version
of USPTO that contains only single-product reactions. Best results
are highlighted in bold. $^{\diamondsuit}$: With beam search (beam
width = 20), $^{\spadesuit}$: Invalid product removal, $^{\clubsuit}$:
Duplicated product removal. \label{tab:reaction_prediction}}
\end{table}
This experiment validates $\Model$ on full reaction product prediction
against the recent state-of-the-art methods \cite{jin2017predicting,schwaller2017found}
using the accuracy metric. The recent method ELECTRO \cite{bradshaw2018predicting}
is not compatible here because it was only evaluated on a subset of
\emph{USPTO} limited to linear chain topology. Comparison against
ELECTRO is reported separately in Appendix~\ref{subsec:Comparison-With-ELECTRO}.
Table~\ref{tab:reaction_prediction} shows the prediction results.
We produce multiple reaction product candidates by using beam search
decoding with beam width $N=20$. Details about beam search and its
behaviors are presented in Appendix~\ref{subsec:Beam-Search}.

In brief, we compute the normalized-over-length log probabilities
of $N$ predicted sequences of reaction triples and sort these values
in descending order to get a rank list of $N$ possible reaction outcomes.
Given a predicted sequence of reaction triples $(u,v,b)^{0:T}$, we
can generate reaction products from input reactants simply by replacing
the old bond of $(u,v)^{\tau}$ with $b^{\tau}$. However, these products
are not guaranteed to be valid (e.g., maximum valence constraint violation
or aromatic molecules cannot be kekulized) so we post-process the
outputs by removing all invalid products. The removal increases the
top-1 accuracy by about 8\% and 10\% on \emph{USPTO-15k} and \emph{USPTO},
respectively. Due to the permutation invariance of the predicted sequence
of reaction triples, some product candidates are duplicate and will
also be removed. This does not lead to any change in \emph{P@1} but
slightly improves \emph{P@3} and \emph{P@5} by about 0.5-1\% on the
two datasets.

Overall, $\Model$ with beam search and post-processing outperforms
both WLDN \cite{jin2017predicting} and Seq2Seq \cite{schwaller2017found}
in the top-1 accuracy. For the top-3 and top-5, our model's performance
is comparable to WLDN's on \emph{USPTO-15k} and is worse than WLDN's
on \emph{USPTO}. It is not surprising since our model is trained to
accurately predict the top-1 outcomes instead of ranking the candidates
directly like WLDN. It is important to emphasize that we did not tune
the model hyper-parameters when training on \emph{USPTO} but reused
the optimal settings from \emph{USPTO-15k} (which is 25 times smaller
than \emph{USPTO}) so the results may not be optimal (see Appendix~\ref{subsec:Model-Configurations}
for more training detail).

\section{Related Work\label{sec:Related-Work}}

\subsection{Learning to Predict Chemical Reaction}

In chemical reaction prediction, machine learning has replaced rule-based
methods \cite{chen2009no} for better generalizability and scalability.
Existing machine learning-based techiques are either template-free
\cite{kayala2011machine,jin2017predicting,fooshee2018deep} and template-based
\cite{wei2016neural,segler2017neural,coley2017prediction}. Both groups
share the same mechanism: running multiple stages with the aid of
reaction templates or rules. For example, in \cite{wei2016neural}
the authors proposed a two-stage model that first classifies reactions
into different types based on the neural fingerprint vectors \cite{duvenaud2015convolutional}
of reactant and reagent molecules. Then, it applies pre-designed SMARTS
transformation on the reactants with respect to the most suitable
predicted reaction type to generate the reaction products. 

The work of \cite{jin2017predicting} treats a reaction as a set of
bond changes so in the first step, they predict which atom pairs are
likely to be reactive using a variant of graph neural networks called
Weisfeiler-Lehman Networks (WLNs). In the next step, they do almost
the same as \cite{coley2017prediction} by modifying the bond type
between the selected atom pairs (with chemical rules satisfied) to
create product candidates and rank them (with reactant molecules as
addition input) using another kind of WLNs called Weifeiler-Lehman
Different Networks (WLDNs).

To the best of our knowledge, \cite{jin2017predicting} is the first
work that achieves remarkable results (with the Precision@1 is about
79.6\%) on the large \emph{USPTO} dataset containing more than 480
thousands reactions. Works of \cite{nam2016linking} and \cite{schwaller2017found}
avoid multi-stage prediction by building a seq2seq model that generates
the (canonical) SMILES string of the single product from the concatenated
SMILES strings of the reactants and reagents in an end-to-end manner.
However, their methods cannot deal with sets of reactants/reagents/products
properly as well as cannot provide concrete reaction mechanism for
every reaction.

The most recent work on this topic is \cite{bradshaw2018predicting}
which solves the reaction prediction problem by predicting a sequence
of bond changes given input reactants and reagents represented as
graphs. To handle ordering, they only select reactions with predefined
topology. Our method, by contrast, is order-free and can be applied
to almost any kind of reactions.

\subsection{Graph Neural Networks for Modeling Molecules}

In recent years, there has been a fast development of graph neural
networks (GNNs) for modeling molecules. These models are proposed
to solve different problems in chemistry including toxicity prediction
\cite{duvenaud2015convolutional}, drug activity classification \cite{shervashidze2011weisfeiler,dai2016discriminative,pham2018graph},
protein interface prediction \cite{fout2017protein} and drug generation
\cite{simonovsky2018graphvae,jin2018junction}. Most of them can be
regarded as variants of message-passing graph neural networks (MPGNNs)
\cite{gilmer2017neural}.

\subsection{Reinforcement Learning for Structural Reasoning}

Reinforcement learning (RL) has become a standard approach to many
structural reasoning problems\footnote{Structural reasoning is a problem of inferring or generating new structure
(e.g. objects with relations)} because it allows agents to perform discrete actions. A typical example
of using RL for structural reasoning is drug generation \cite{li2018multi,you2018graph}.
Both \cite{li2018multi} and \cite{you2018graph} learn the same generative
policy whose action set including: i) adding a new atom or a molecular
scaffold to the intermediate graph, ii) connecting existing pair of
atoms with bonds, and iii) terminating generation. However, \cite{you2018graph}
uses an adversarial loss to enforce global chemical constraints on
the generated molecules as a whole instead of using the common reconstruction
loss as in \cite{li2018multi}. Other examples are path-based relational
reasoning in knowledge graphs \cite{das2017go} and learning combinatorial
optimization over graphs \cite{dai2017learning}.

\section{Discussion\label{sec:Discussion}}

We have introduced a novel method named Graph Transformation Policy
Network ($\Model$) for predicting products of a chemical reaction.
$\Model$ uses graph neural networks to represent input reactant and
reagent molecules, and uses reinforcement learning to find an optimal
sequence of bond changes that transforms the reactants into products.
We train $\Model$ using the Advantage Actor-Critic (A2C) method with
appropriate constraints to account for notable aspects of chemical
reaction. Experiments on real datasets have demonstrated the competitiveness
of our model.

Although the $\Model$ was proposed to solve the chemical reaction
problem, it is indeed generic to solve the graph transformation problem,
which can be useful in reasoning about relations (e.g., see \cite{zambaldi2018relational})
and changes in relation. Open rooms include addressing dynamic graphs
over time, extending toward full chemical planning and structural
reasoning using RL.


\appendix
\newpage{}

\section{Appendix\label{sec:Appendix}}

\global\long\def\signal{\text{signal}}%
\global\long\def\atp{\text{atom pair}}%
\global\long\def\bond{\text{bond}}%
\global\long\def\extract{\text{extract}}%

\subsection{Graph Neural Network\label{subsec:Message-Passing-Graph}}

In this section, we describe our graph neural network (GNN) in detail.
Since our GNN does not use the recurrent hidden state $\hb^{\tau}$,
we exclude the time step $\tau$ from our notations for clarity. Instead,
we use $t$ to denote a message passing step.

\subsubsection*{Graph notations}

Input to our GNN is a graph $\Graph=(\Vertices,\Edges)$ in which
each node $i\in\Vertices$ is represented by a node feature vector
$\vb_{i}$ and each edge $(i,j)\in\Edges$ is represented by an edge
feature vector $\eb_{ij}$. For example of molecular graph, the node
feature vector $\vb_{i}$ may include chemical information about the
atom $i$ such as its type, charge and degree. Similarly, $\eb_{ij}$
captures the bond type between the two atoms $i$ and $j$. We denote
by $\Neigh(i)$ the set of all neighbor nodes of node $i$ together
with their links to node $i$:

\begin{eqnarray*}
\Neigh(i) & \equiv & \left\{ \left(j,e_{ij}\right)\mid j\ \text{is a neighbor node of}\ i\right\} 
\end{eqnarray*}
If we only care about the neighbor nodes of $i$ not their links,
we use the notation $\Neigh_{\text{n}}(i)$ defined as:

\begin{eqnarray*}
\Neigh_{\text{n}}(i) & \equiv & \left\{ j\mid j\ \text{is a neighbor node of}\ i\right\} 
\end{eqnarray*}
In addition to $\vb_{i}$, node $i$ also has a state vector $\xb_{i}$
to store information about itself and the surrounding context. This
state vector is updated recursively using the neural message passing
method \cite{battaglia2016interaction,pham2017column,hamilton2017inductive,gilmer2017neural,schlichtkrull2017modeling}.
The initial state $\xb_{i}^{0}$ is the nonlinear mapping of $\vb_{i}$:
\begin{eqnarray}
\xb_{i}^{0} & = & \sigma\left(W\vb_{i}+b\right)\label{eq:init_node_state}
\end{eqnarray}

\subsubsection*{Computing neighbor messages}

At the message passing step $t$, we compute the message $\mb_{ij}^{t}$
from every neighbor node $j\in\Neigh_{n}(i)$ to node $i$ as:

\begin{eqnarray}
\mb_{ij}^{t} & = & f\left(\xb_{i}^{t},\xb_{j}^{t},\eb_{ij}\right)\nonumber \\
 & = & \sigma\left(W\left[\xb_{i}^{t},\xb_{j}^{t},\eb_{ij}\right]+b\right)\label{eq:compute_neigh_msn}
\end{eqnarray}
where $\left[\cdot\right]$ denotes concatenation; and $\sigma$ is
a nonlinear function.

\subsubsection*{Aggregating neighbor messages}

Then, we aggregate all the messages sent to node $i$ into a single
message vector by averaging:

\begin{eqnarray}
\mb_{i}^{t} & = & \frac{1}{\left|\Neigh_{\text{n}}(i)\right|}\sum_{j\in\Neigh_{\text{n}}(i)}\mb_{ij}^{t}\label{eq:aggr_neigh_msn}
\end{eqnarray}
where $|\Neigh_{\text{n}}(i)|$ is the number of neighbor nodes of
node $i$. 

\subsubsection*{Updating node state}

Finally, we update the state of node $i$ as follows:

\begin{eqnarray}
\xb_{i}^{t+1} & = & g\left(\xb_{i}^{t},\mb_{i}^{t},\vb_{i}\right)\label{eq:generic_node_update}
\end{eqnarray}
where $g(.)$ is a Highway Network \cite{srivastava2015training}:

\begin{eqnarray}
\xb_{i}^{t+1} & = & \text{Highway}\left(\xb_{i}^{t},\mb_{i}^{t},\vb_{i}\right)\label{eq:highway_node_update}\\
 & = & \alphab\ast\tilde{\xb}_{i}^{t+1}+(1-\alphab)\ast\xb_{i}^{t}\label{eq:gating_mechanism}
\end{eqnarray}
where $\tilde{\xb}_{i}^{t+1}$ is the nonlinear part which is computed
as:
\begin{align*}
\bar{\xb}_{i}^{t+1} & =\sigma\left(W_{1}\left[\xb_{i}^{t},\mb_{i}^{t},\vb_{i}^{t}\right]+b_{1}\right)
\end{align*}
and $\alphab$ is the gate controlling the flow of information:

\[
\alphab=\text{sigmoid}(W_{2}\left[\xb_{i}^{t},\mb_{i}^{t},\vb_{i}^{t}\right]+b_{2})
\]
By combining Eqs.~(\ref{eq:compute_neigh_msn},\ref{eq:aggr_neigh_msn},\ref{eq:highway_node_update})
together, one step of message passing update for node $i$ can be
written in a generic way as follows:

\begin{equation}
\xb_{i}^{t+1}=\text{MessagePassing}\left(\xb_{i}^{t},\vb_{i},\Neigh(i)\right)\label{eq:MsgPass}
\end{equation}

\subsection{Updating States\label{subsec:Updating-States}}

\subsubsection*{Updating RNN state}

We keep the old representation of the edge that have been modified
in the hidden memory of the RNN as follows: 

\begin{equation}
\hb^{\tau}=\text{GRU}\left(\hb^{\tau-1},\zb_{uv}^{\tau}\right)\label{eq:RNN_update}
\end{equation}
where $\text{GRU}$ stands for Gated Recurrent Units \cite{cho2014learning};
$\zb_{uv}^{\tau}$ is the representation vector of the atom pair $(u,v)^{\tau}$
including its old bond (see Eq.~\ref{eq:np_hid_global}). Eq.~(\ref{eq:RNN_update})
allows the model to keep track of all the changes happening to the
graph so far so it can make more accurate prediction later.

\subsubsection*{Updating graph structure and node states}

After predicting a reaction triple $(u,v,b)^{\tau}$ at step $\tau$,
we update the graph structure and node states based on the new bond
change. First, to update the graph structure, we simply update the
neighbor set of $u$ and $v$ with information from the other atom
and the new bond type $b$ as follows:

\begin{eqnarray}
\Neigh^{\tau}(u) & = & \left(\Neigh^{\tau-1}(u)\backslash\left(v,b^{\text{old}}\right)\right)\cup\left(v,b\right)\label{eq:u_neigh_update}\\
\Neigh^{\tau}(v) & = & \left(\Neigh^{\tau-1}(v)\backslash\left(u,b^{\text{old}}\right)\right)\cup\left(u,b\right)\label{eq:v_neigh_update}
\end{eqnarray}
Next, to update the node states, our model performs one step of message
passing for $u$ and $v$ with their new neighbor sets: 

\begin{eqnarray}
\xb_{u}^{\tau} & = & \text{MessagePassing}\left(\xb_{u}^{\tau-1},\vb_{u},\Neigh^{\tau}(u)\right)\label{eq:u_state_update}\\
\xb_{v}^{\tau} & = & \text{MessagePassing}\left(\xb_{v}^{\tau-1},\vb_{v},\Neigh^{\tau}(v)\right)\label{eq:v_state_update}
\end{eqnarray}
where the $\text{MessagePassing}(.)$ function is defined in Eq.~(\ref{eq:MsgPass}).
For other nodes in the graph to be aware of the new structures of
$u$ and $v$, we need to perform several message passing steps for
all nodes in the graph after Eqs.~(\ref{eq:u_state_update}, \ref{eq:v_state_update}).
However, it is very costly to run for every prediction step $\tau$.
Sometimes it is unnecessary since far-away bonds are less likely to
be affected by the current bond change (unless the far-way bonds and
the new bond are in an aromatic ring). Therefore, in our model, we
limit the number of message passing updates for all nodes at step
$\tau$ to be $1$.

\subsection{Model Configurations\label{subsec:Model-Configurations}}

We optimize our model's hyper-parameters in two stages: First, we
tune the hyper-parameters of the GNN and the NPPN for the reaction
atom pair prediction task. Then, we fix the optimal settings of the
first two components and optimize the hyper-parameters of the PN for
the reaction product prediction task. 

We provide details about the settings that give good results on the
\emph{USPTO-15k} dataset below. With these settings, we trained another
model on the \emph{USPTO} dataset from scratch. Because training on
the large dataset such as the \emph{USPTO} takes time, we did not
tune hyper-parameters on the \emph{USPTO}, eventhough it is possible
to increase model sizes for better performance.

Unless explicitly stated, all neural networks in our model have 2
layers with the same number of hidden units, ReLU activation and residual
connections \cite{he2016deep}.

\paragraph*{Graph Neural Network (GNN)}

There are 72 different types of atom depending on their atomic numbers
and 5 different types of bond including NULL, SINGLE, DOUBLE, TRIPLE
and AROMATIC. The size of embedding vectors for atom and bond are
51 and 21, respectively. Apart from atom type, each atom has 5 more
attributes listed in Table~\ref{tab:atom_features}. These attributes
are normalized to the range of {[}0, 1{]} and are concatenated to
the atom embedding vector to form a final atom feature vector of size
56. The state vector and the neighbor message vector for an atom both
have the size of 99. The number of message passing steps is 6.

\begin{table}
\begin{centering}
\begin{tabular}{|l|c|}
\hline 
Atom attribute & Data type\tabularnewline
\hline 
\hline 
Degree & numeric\tabularnewline
\hline 
Explicit valence & numeric\tabularnewline
\hline 
Explicit number of Hs & numeric\tabularnewline
\hline 
Charge & numeric\tabularnewline
\hline 
Part of a ring & boolean\tabularnewline
\hline 
\end{tabular}
\par\end{centering}
\caption{Data types of atom attributes. \label{tab:atom_features}}
\end{table}

\paragraph*{Node Pair Prediction Network (NPPN)}

This component consists of two parts. The first part computes the
representation vector $\zb_{ij}$ of an atom pair $(i,j)$ using a
neural network with hidden size of 71. The second part maps $\zb_{ij}$
to an unnormalized score $s_{ij}$ using the function $f^{\atp}$
(see Eqs.~(\ref{eq:np_score_local},\ref{eq:np_score_global})).
This function is also a neural network with hidden size of 51.

\paragraph*{Policy Network (PN)}

The recurrent network is a GRU \cite{cho2014learning} with 101 hidden
units. The value function $V_{\phi}$ is a neural network with 99
hidden units. The two functions $f^{\signal}$ for computing signal
scores (see Eq.~(\ref{eq:signal_prob})) and $f^{\bond}$ for computing
scores over bond types (see Eq.~(\ref{eq:bond_prob})) are neural
networks with 81 hidden units.

\paragraph*{Training}

At each step, we set the reward to be 1.0 for correct prediction of
signal/atom pair/bond type and -1.0 for incorrect prediction. After
the prediction sequence is terminated (zero signal was emitted), we
check whether the entire set of predicted reaction triples is correct
or not. If it is correct, we give the model a reward value of 2.0,
otherwise -2.0. From the rewards and estimated values for signal,
atom pair and bond type, we define the Advantage Actor Critic loss
(A2C) as in Eq.~(\ref{eq:actor_critic_loss3}). The coefficients
of components in the final loss $\Loss$ are set empirically as follows:

\[
\Loss=\Loss^{\text{A2C}}+0.5\times\Loss^{\text{value}}+\Loss^{\text{atom pair}}+0.2\times\Loss^{\text{over length}}+0.2\times\Loss^{\text{in top }K}
\]
We trained our model using Adam \cite{kingma2014adam} with the initial
learning rate of 0.001 for both \emph{USPTO-15k} and \emph{USPTO}.
For \emph{USPTO-15k}, the learning rate will decrease by half if the
\emph{Precision@1} does not improve on the validation set after 1,000
steps until it reaches the minimum value of $5\times10^{-5}$. For
\emph{USPTO}, the decay rate is 0.8 after every 500 steps of no improvement
until reaching the minimum learning rate is $2\times10^{-5}$. The
maximum number of training iterations is $10^{6}$ and the batch size
is 20.

\subsection{Decoding with Beam Search\label{subsec:Beam-Search}}

For decoding, our model generates a sequence of reaction triples (including
the stop signal) $(\xi,u,v,b)$ by taking the best $(u,v)$ and $b$
at every step until it outputs a zero signal ($\xi=0$). In other
words, it computes the argmax of $p\left((\xi,u,v,b)^{\tau}\mid\Graph,(\xi,u,v,b)^{0:\tau-1}\right)$
at every step $\tau$. However, this algorithm is not robust for the
sequence generation task because just a single error at a step may
destroy the entire sequence. To overcome this issue, we employ beam
search for decoding.

During beam search, we keep track of $N>1$ best subsequences at every
step $\tau$. $N$ is called beam width. Instead of modeling the conditional
distribution of generating an output at the current step $\tau$,
we model the joint distribution of the whole subsequence that has
been generated from $0$ to $\tau$:

\begin{eqnarray}
\log p\left((\xi,u,v,b)^{0:\tau}|\Graph\right) & = & \log p\left((\xi,u,v,b)^{\tau}|\Graph,(\xi,u,v,b)^{0:\tau-1}\right)+\nonumber \\
 &  & \log p\left((\xi,u,v,b)^{0:\tau-1}|\Graph\right)\label{eq:joint_seq_prob}
\end{eqnarray}
Computing all configurations of $(\xi,u,v,b)^{\tau}$ jointly is
very memory demanding, however. Thus, we decompose the first term
as follows:

\begin{eqnarray*}
\log p\left((\xi,u,v,b)^{\tau}|\Graph,(\xi,u,v,b)^{0:\tau-1}\right) & = & \log p\left(\xi^{\tau}|\Graph,(\xi,u,v,b)^{0:\tau-1}\right)+\\
 &  & \log p\left((u,v)^{\tau}|\xi^{\tau},\Graph,(\xi,u,v,b)^{0:\tau-1}\right)+\\
 &  & \log\left(b^{\tau}|(\xi,u,v)^{\tau},\Graph,(\xi,u,v,b)^{0:\tau-1}\right)
\end{eqnarray*}
At step $\tau$, we do beam search for the signal $\xi^{\tau}$, then
the atom pair $(u,v)^{\tau}$ and finally the bond type $b^{\tau}$.
Algorithm~\ref{alg:beam_search} describes beam search in detail.
Some notable technicalities are:
\begin{itemize}
\item We only do beam search for $(u,v)$ and $b$ if the prediction is
ongoing, i.e., when $\xi^{\tau}=1$. To keep track of this, we use
a boolean vector $C$ of length $N$ with $C^{0}$ is initialized
to be all true.
\item To avoid beam search favoring short sequences, we normalize the log
probability scores over sequence lengths. This is shown in lines \ref{length_norm_1},
\ref{length_norm_2}, \ref{length_norm_3} and \ref{length_norm_4}
\end{itemize}
\begin{algorithm}
\begin{algorithmic}[1]

\Require{A multi-graph $\Graph$ consisting of reactant and reagent
molecules, number of bond types $E$, max prediction steps $T$, beam
width $N$}

\State{$P^{0}=\left[(-1,-1,-1,-1),...\right]$}\Comment{The best
$N$ subsequences of $(\xi,u,v,b)$}

\State{$S^{0}=\left[0,...\right]$}\Comment{The length-normalized
log joint probabilities of the best $N$ subsequences}

\State{$C^{0}=\left[\text{True},...\right]$}\Comment{The continuation
indicator of the best $N$ subsequences}

\Statex{}

\State{Perform $L$ steps of message passing for all nodes using
Eq.~(\ref{eq:atom_repr_vector})}

\State{$\xb_{i}^{0}=\xb_{i}\ \forall i\in\Vertices$}\Comment{The
initial states of all nodes before decoding}

\State{$\Neigh^{0}(i)=\Neigh(i)\ \forall i\in\Vertices$}\Comment{The
initial neighbor set of all nodes before decoding}

\State{$\hb^{0}$ is loaded from the saved model}\Comment{The initial
RNN hidden state before decoding}

\Statex{}

\For{$\tau$ from $1$ to $T$ }

\State{Find the top $K$ atom pairs $\left\{ (u_{k},v_{k})^{\tau}\mid k=\overline{1,K}\right\} $
using Eqs.~(\ref{eq:np_hid_global},\ref{eq:np_score_global})}

\Statex{}

\State{$S^{\ensuremath{\tau-1;0}}=S^{\tau-1}\times\frac{\tau-1}{\tau}$}\Comment{Superscript
$0$ denotes the sub-step 0}\label{length_norm_1}

\State{$P^{\tau-1;0}=P^{\tau-1}$; $C^{\tau-1;0}=C^{\tau-1}$}

\Statex{}

\State{\textcolor{gray}{\textit{Beam search for continuation signals}}}

\State{\textcolor{gray}{\rule[0.5ex]{0.5\textwidth}{1pt}}}

\State{$R^{\signal}=\emptyset$}\Comment{Stores the log joint probabilities
for $N\times2$ possible signals}

\For{$n$ from $1$ to $N$}

\State{Compute $p\left(\xi^{\tau}\mid P_{n}^{\tau-1;0}\right)$ using
Eq.~(\ref{eq:signal_prob})}

\State{Add $C^{\tau-1;0}\times\frac{1}{\tau}\log p\left(\xi^{\tau}=\delta\mid P_{n}^{\tau-1;0}\right)+S_{n}^{\tau-1;0}$
to $R^{\signal}$ for $\delta\in\{\text{True},\text{False}\}$}\label{length_norm_2}

\EndFor

\State{Sort $R^{\signal}$ in descending order}

\State{$S^{\ensuremath{\tau-1;1}}=R_{0:N}^{\signal}$}

\State{$\bar{\xi}^{\tau}\equiv$ output signal of $N$ beams in $R_{0:N}^{\signal}$}

\State{$I^{\tau-1;1}\equiv$ indices of $N$ beams in $R_{0:N}^{\signal}$
}

\State{$P^{\tau-1;1}=$ $\extract\left(P^{\tau-1;0},I^{\tau-1;1}\right)$}

\State{$C^{\tau-1;1}=\extract\left(C^{\tau-1;0},I^{\tau-1;1}\right)$}

\State{$C_{n}^{\tau-1;1}=C_{n}^{\tau-1;1}\wedge\xi_{n}^{\tau}$ $\forall n\in\overline{1,N}$
}

\State{\textcolor{gray}{\rule[0.5ex]{0.5\textwidth}{1pt}}}

\Statex{}

\State{\textcolor{gray}{\textit{Beam search for atom pairs}}}

\State{\textcolor{gray}{\rule[0.5ex]{0.5\textwidth}{1pt}}}

\State{$R^{\atp}=\emptyset$}\Comment{Stores the log joint probabilities
for $N\times K$ possible atom pairs}

\For{$n$ from $1$ to $N$}

\State{Compute $p\left((u,v)^{\tau}|\bar{\xi}_{n}^{\tau},P_{n}^{\tau-1;1}\right)$
using Eq.~(\ref{eq:atom_pair_prob})}

\State{Add $C^{\tau-1;1}\times\frac{1}{\tau}\log p\left((u,v)_{k}^{\tau}|\xi_{n}^{\tau},P_{n}^{\tau-1;1}\right)+S_{n}^{\tau-1;1}$
to $R^{\atp}$ $\forall k\in\overline{1,K}$}\label{length_norm_3}

\EndFor

\State{Sort $R^{\atp}$ in descending order}

\State{$S^{\ensuremath{\tau-1;2}}=R_{0:N}^{\atp}$}

\State{$(\bar{u},\bar{v})^{\tau}\equiv$ output atom pair of $N$
beams in $R_{0:N}^{\atp}$}

\State{$I^{\tau-1;2}\equiv$ indices of $N$ beams in $R_{0:N}^{\atp}$}

\State{$P^{\tau-1;2}=\extract\left(P^{\tau-1;1},I^{\tau-1;2}\right)$}

\State{$C^{\tau-1;2}=\extract\left(C^{\tau-1;1},I^{\tau-1;2}\right)$}

\State{$\bar{\xi}^{\tau}=\extract\left(\bar{\xi}^{\tau},I^{\tau-1;2}\right)$}

\State{\textcolor{gray}{\rule[0.5ex]{0.5\textwidth}{1pt}}}

\Statex{}

\State{\textcolor{gray}{\textit{Beam search for bonds}}}

\State{\textcolor{gray}{\rule[0.5ex]{0.5\textwidth}{1pt}}}

\State{$R^{\bond}=\emptyset$}\Comment{Stores the log joint probabilities
for $N\times B$ possible bonds}

\algstore{mybreak}

\end{algorithmic}

\caption{Reaction triple prediction using beam search.\label{alg:beam_search}}
\end{algorithm}
\begin{algorithm}
\begin{algorithmic}[1]

\algrestore{mybreak}

\For{$n$ from $1$ to $N$}

\State{Compute $p\left(b^{\tau}\mid(\bar{\xi},\bar{u},\bar{v})_{n}^{\tau},P_{n}^{\tau-1}\right)$
using Eq.~(\ref{eq:bond_prob})}

\State{Add $C^{\tau-1;2}\times\frac{1}{\tau}\log p\left(b^{\tau}=\beta\mid(\bar{\xi},\bar{u},\bar{v}),P_{n}^{\tau-1}\right)+S_{b}^{\tau-1}$
to $R^{\bond}$ $\forall\beta\in\overline{1,B}$}\label{length_norm_4}

\EndFor

\State{Sort $R^{\bond}$ in descending order}

\State{$S^{\ensuremath{\tau-1;3}}=R_{0:N}^{\bond}$}

\State{$\bar{b}^{\tau}\equiv$ output bond of $N$ beams in $R_{0:N}^{\bond}$}

\State{$I^{\tau-1;3}\equiv$ indices of $N$ beams in $R_{0:N}^{\bond}$
}

\State{$P^{\tau-1;3}=\extract\left(P^{\tau-1;2},I^{\tau-1;3}\right)$}

\State{$C^{\tau-1;3}=\extract\left(C^{\tau-1;2},I^{\tau-1;3}\right)$}

\State{$\bar{\xi}^{\tau}=\extract\left(\bar{\xi}^{\tau},I^{\tau-1;3}\right)$}

\State{$(\bar{u},\bar{v})^{\tau}=\extract\left((\bar{u},\bar{v})^{\tau},I^{\tau-1;3}\right)$}

\State{\textcolor{gray}{\rule[0.5ex]{0.5\textwidth}{1pt}}}

\Statex{}

\State{$S^{\tau}=S^{\tau-1;3}$; $C^{\tau}=C^{\tau-1;3}$}

\State{$P_{n}^{\tau}=$ append$\left(P_{n}^{\tau-1;3},(\bar{\xi},\bar{u},\bar{v},\bar{b})_{n}^{\tau}\right)$}

\Statex{}

\For{$n$ from $1$ to $N$}

\State{Update the $\Neigh^{\tau}(\bar{u}_{n})$ and $\Neigh^{\tau}(\bar{v}_{n})$
for all $n=\overline{1,N}$ using Eqs.~(\ref{eq:u_neigh_update},\ref{eq:v_neigh_update})}

\State{Update $\xb_{\bar{u}_{n}}^{\tau}$ and $\xb_{\bar{v}_{n}}^{\tau}$
using Eq.~(\ref{eq:atom_repr_vector})}

\State{Perform $m$ steps of message passing for all nodes in the
graph}

\State{Update $\hb^{\tau}$ using Eq.~(\ref{eq:RNN_update})}

\EndFor

\EndFor

\Ensure{$P^{\ensuremath{T}}$, $S^{T}$}

\end{algorithmic}

\caption{Reaction triple prediction using beam search (cont.)}
\end{algorithm}

\subsubsection*{Beam width analysis}

Table~\ref{tab:beam_width_table} reports how beam width affects
the decoding performance on the \emph{USPTO-15k} dataset. Surprisingly,
the top-1 accuracy in case of beam width\footnote{Note that beam search with beam width = 1 is different from greedy
search as in beam search, as we model the whole sequence probability.} of 1 is higher than the those when beam widths range from 2 to 15.
It means that large beam width is not always good in our situation.
However, at beam width of 20, our beam search achieves the best results
for different values of $k$. Thus, we set the beam width to 20 in
subsequent experiments.

\begin{table}
\begin{centering}
\begin{tabular}{|c|c|c|c|c|c|c|c|}
\hline 
\multirow{2}{*}{Beam width} & \multicolumn{7}{c|}{\emph{Precision@k}}\tabularnewline
\cline{2-8} 
 & 1 & 2 & 3 & 5 & 10 & 15 & 20\tabularnewline
\hline 
\hline 
1 & 74.49 & - & - & - & - & - & -\tabularnewline
\hline 
2 & 72.21 & 80.65 & - & - & - & - & -\tabularnewline
\hline 
5 & 72.21 & 79.54 & 82.29 & \textbf{84.27} & - & - & -\tabularnewline
\hline 
10 & 72.15 & 79.54 & 82.19 & 83.93 & 86.01 & - & -\tabularnewline
\hline 
15 & 72.15 & 79.54 & 82.16 & 83.93 & 86.11 & 86.98 & -\tabularnewline
\hline 
20 & \textbf{74.56} & \textbf{80.72} & \textbf{82.62} & 84.23 & \textbf{86.14} & \textbf{87.04} & \textbf{87.55}\tabularnewline
\hline 
\end{tabular}
\par\end{centering}
\caption{Reaction product prediction results using beam search with different
values of beam width on \emph{USPTO-15k}. \label{tab:beam_width_table}}
\end{table}

\subsection{Using Reagent Information Explicitly\label{subsec:Using-Reagents-Information}}

\begin{figure}
\begin{centering}
\includegraphics[width=0.5\textwidth]{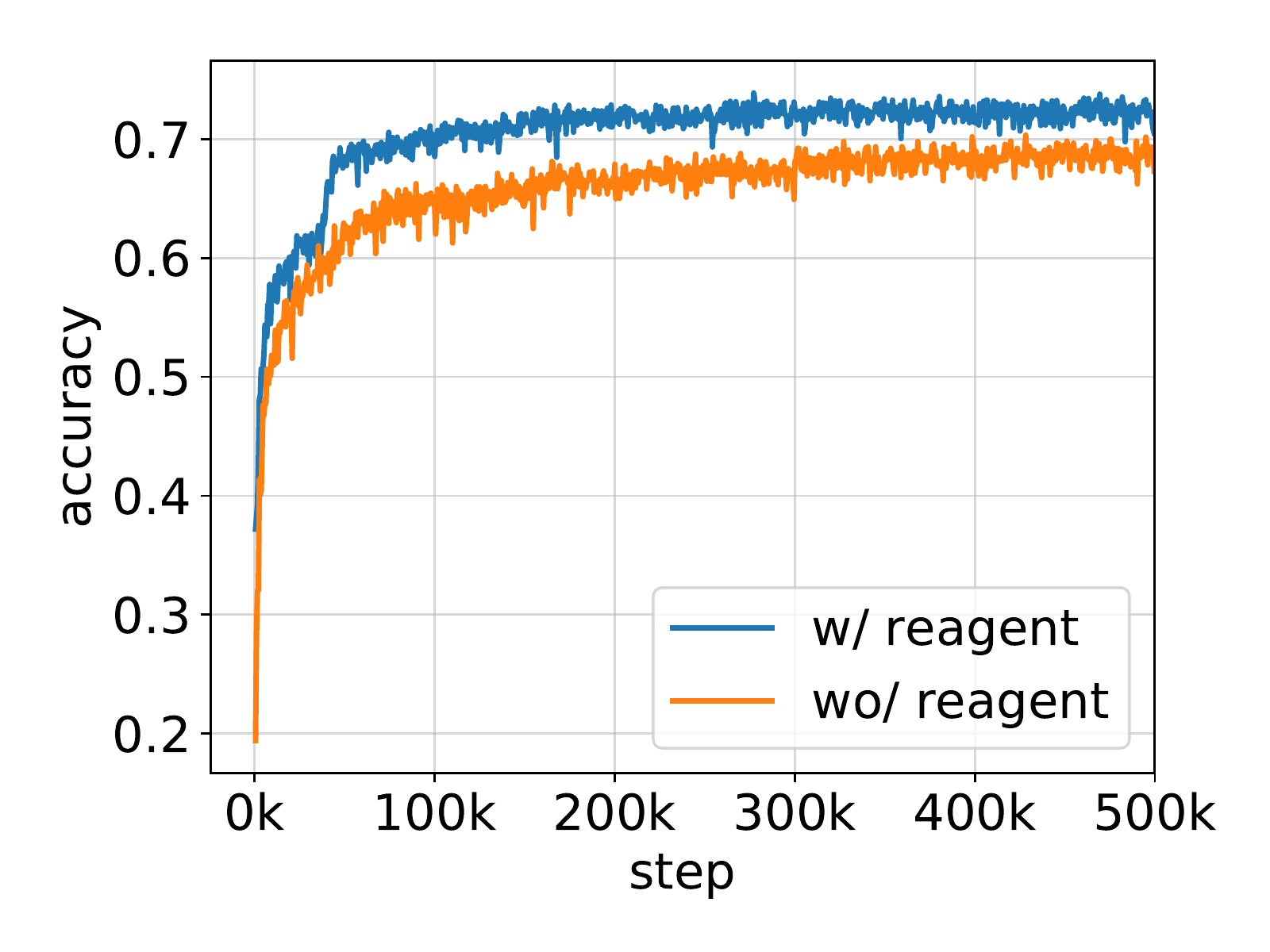}
\par\end{centering}
\caption{Learning curves of our model with and without using reagent information
explicitly on \emph{USPTO-15k}.\label{fig:reagent_learning_curve}}
\end{figure}

As can be seen from Table~\ref{tab:reagent_statistics}, reagent
molecules account for about a half of the input molecules on average
and 60-80\% of all reactions containing reagents. It suggests that
the proper use of reagent information will lead to better prediction.
In our model, before computing the scores for all atom pairs, we append
to the representation vector of every atom a binary scalar indicating
whether this atom comes from a reagent molecule or not. Then, at the
top-$K$ atom pair selection step, we also exclude all atom pairs
that have either atoms belong to a reagent molecule. The improvement
in prediction accuracy on the validation set of \emph{USPTO-15k} is
shown in Fig.~\ref{fig:reagent_learning_curve}.

\begin{table}
\begin{centering}
\begin{tabular}{|c|c|>{\centering}p{0.2\textwidth}|>{\centering}p{0.2\textwidth}|}
\hline 
\multicolumn{2}{|c|}{Dataset} & \%reactions \\
containing reagents & \%reagents over \\
input molecules\tabularnewline
\hline 
\hline 
\multirow{3}{*}{\emph{USPTO-15k}} & train & 63.1\% & 41.3\%\tabularnewline
\cline{2-4} 
 & valid & 65.3\% & 42.3\%\tabularnewline
\cline{2-4} 
 & test & 63.6\% & 40.9\%\tabularnewline
\hline 
\hline 
\multirow{3}{*}{\emph{USPTO}} & train & 79.7\% & 54.0\%\tabularnewline
\cline{2-4} 
 & valid & 80.0\% & 54.4\%\tabularnewline
\cline{2-4} 
 & test & 79.9\% & 54.2\%\tabularnewline
\hline 
\end{tabular}
\par\end{centering}
\caption{Proportion of reactions containing reagents and proportion of reagents
over input molecules on \emph{USPTO-15k} and \emph{USPTO}.\label{tab:reagent_statistics}}
\end{table}

\subsection{Comparison with ELECTRO\label{subsec:Comparison-With-ELECTRO}}

\paragraph{In method}

Both $\Model$ and ELECTRO \cite{bradshaw2018predicting} are able
to explain the mechanism behind a reaction. ELECTRO regards a reaction
as an ordered sequence that alternates between removing and adding
a single bond. Our model, on the other hand, assumes no specific order
of transformations as well as the amount of valences that a bond can
change. Thus, our model is more generic than ELECTRO and can cover
a much larger set of reactions.

\paragraph{In performance}

To do a fair comparison with ELECTRO \cite{bradshaw2018predicting},
we follow their procedure described in the paper to prepare a new
test set that contains only reactions with linear chain topology and
single-valence bond changes. It results in 29,808 reactions, close
to the reported number of 29,360 in \cite{bradshaw2018predicting}.
We reuse our old model (see Section~\ref{subsec:Reaction-Product-Prediction})
trained on the original \emph{USPTO} dataset. We also use beam search
decoding and post-processing as similar to \cite{bradshaw2018predicting}.
From Table~\ref{tab:ELECTRO_comparision}, we see that $\Model$
achieves the highest top-1 accuracy of 87.35\%, outperforming ELECTRO
and WLDN by 0.35\% and 3\%, respectively. For the top-3 and top-5
accuracies, our model, however, does worse than the other two. Especially,
while both ELECTRO and WLDN have big jumps from \emph{P@1} to \emph{P@3}
with about 7\% improvement, $\Model$ only has 3\% increase. We conjecture
that this problem mainly comes from the fact that $\Model$ was not
optimized on the compatible training and validation sets.

\begin{table}
\begin{centering}
\begin{tabular}{|c|c|c|c|}
\hline 
\multirow{2}{*}{Model} & \multicolumn{3}{c|}{\emph{Processed USPTO}}\tabularnewline
\cline{2-4} 
 & \emph{P@1} & \emph{P@3} & \emph{P@5}\tabularnewline
\hline 
\hline 
WLDN \cite{jin2017predicting} & 84.0 & 91.1 & 92.3\tabularnewline
\hline 
ELECTRO \cite{bradshaw2018predicting} & 87.0 & \textbf{94.5} & \textbf{95.9}\tabularnewline
\hline 
\hline 
$\Model$$^{\diamondsuit\spadesuit\clubsuit}$ & \textbf{87.35} & 90.22 & 90.68\tabularnewline
\hline 
\end{tabular}
\par\end{centering}
\caption{Results for the reaction prediction task. \emph{P@k} is the precision
at $k$. Best results are highlighted in bold. Meanings of markers
in our model: $^{\diamondsuit}$: With beam search (beam width = 20),
$^{\spadesuit}$: Invalid product removal, $^{\clubsuit}$: Duplicate
product removal.\label{tab:ELECTRO_comparision}}
\end{table}

\subsection{Error Analysis}

In this section, we analyze several error types that our model makes
during prediction. All the results below are computed on the \emph{USPTO-15k}
dataset by using beam search decoding with the beam width $N=20$
and no post-processing. 

\paragraph{Errors grouped by number of bond changes}

\begin{figure}
\begin{centering}
\subfloat[\label{fig:hit_rate_by_length}]{\begin{centering}
\includegraphics[width=0.48\textwidth]{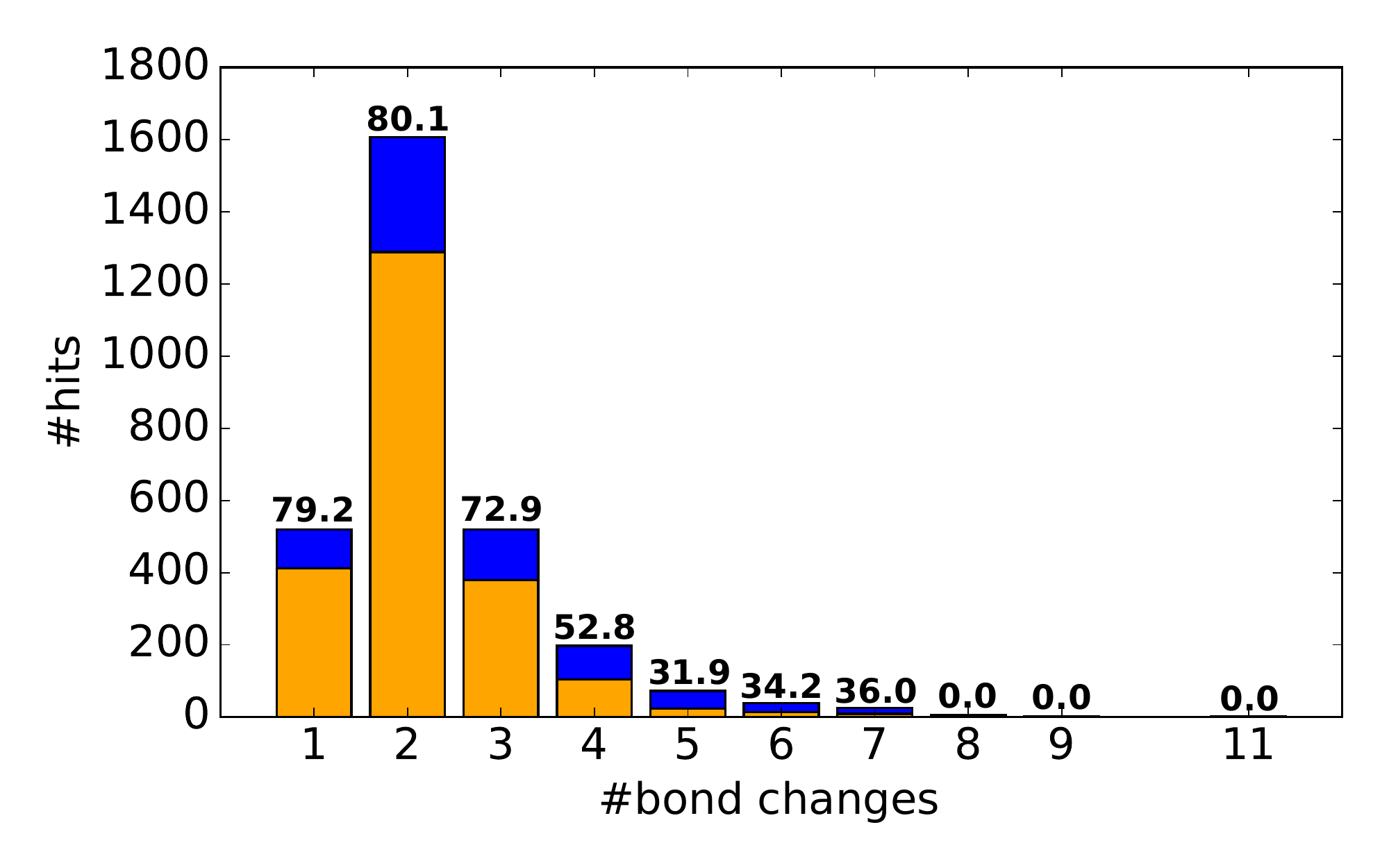}
\par\end{centering}
}\subfloat[\label{fig:error_by_length}]{\begin{centering}
\includegraphics[width=0.48\textwidth]{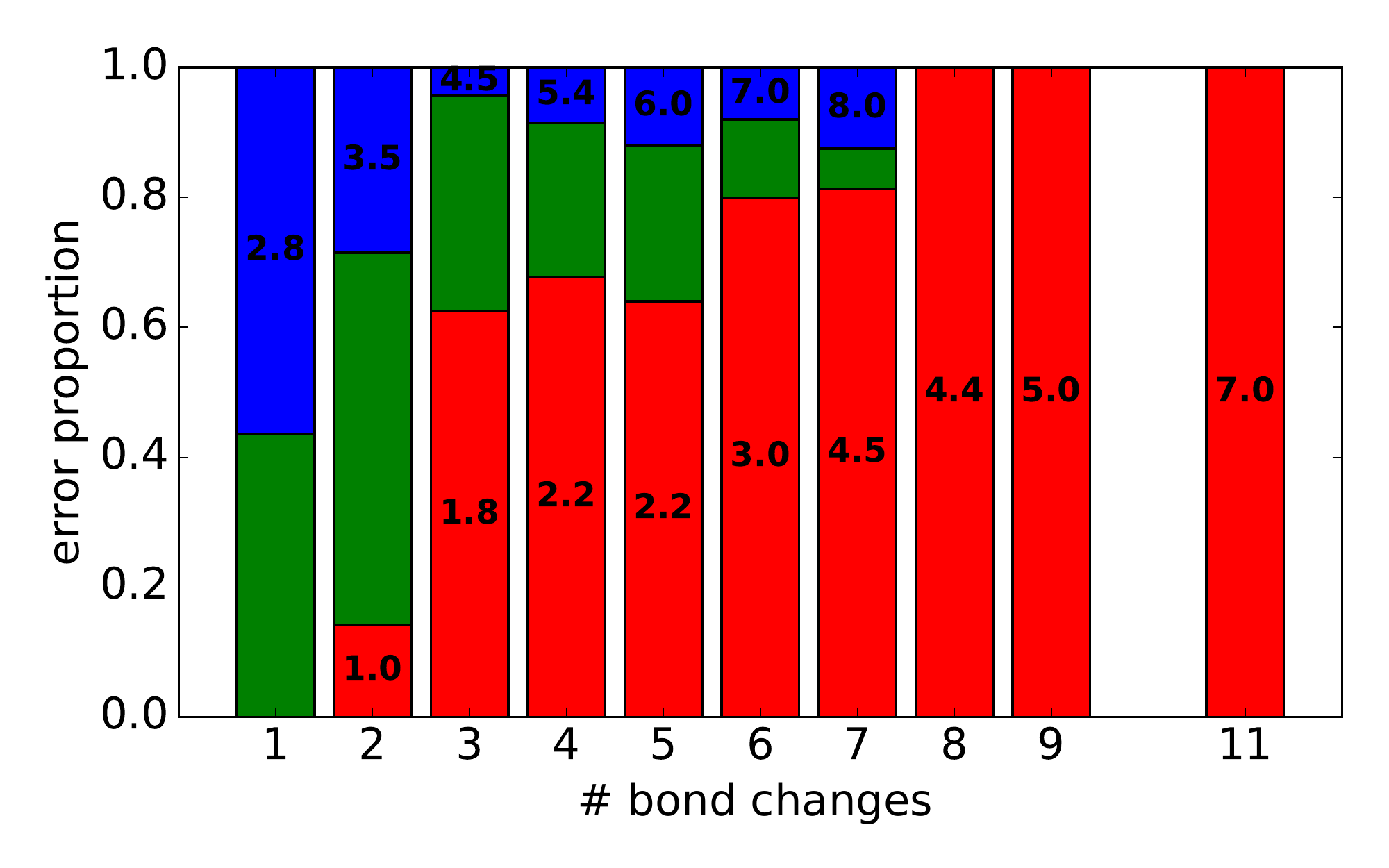}
\par\end{centering}
}
\par\end{centering}
\caption{Performance with respect to different numbers of bond changes. (a)
Top-1 accuracy. (b) Errors grouped by length. In (a), \textbf{blue}:
all reactions having that sequence length; \textbf{orange}: correct
predicted reactions. In (b), \textbf{red}: the predicted sequence
is shorter (than the groundtruth sequence); \textbf{green}: the predicted
and the groundtruth have the same length; \textbf{blue}: the predicted
sequence is longer; number indidate the average length.}
\end{figure}

Fig.~\ref{tab:beam_width_table} shows the top-1 accuracies for reactions
with different number of bond changes. Our model performs poorly on
reactions with many bond changes. However, those kinds of reactions
only accounts for a small proportion in the dataset. From Fig.~\ref{fig:error_by_length},
we see that the lengths of the error sequences tends to be shorter
than the lengths of the groundtruth sequences.

\paragraph{Errors caused by signal/atom pair/bond type}

\begin{figure}
\begin{centering}
\subfloat[Proportion of the first incorrect sub-action that our model makes.\label{fig:error_subaction}]{\begin{centering}
\includegraphics[width=0.4\textwidth]{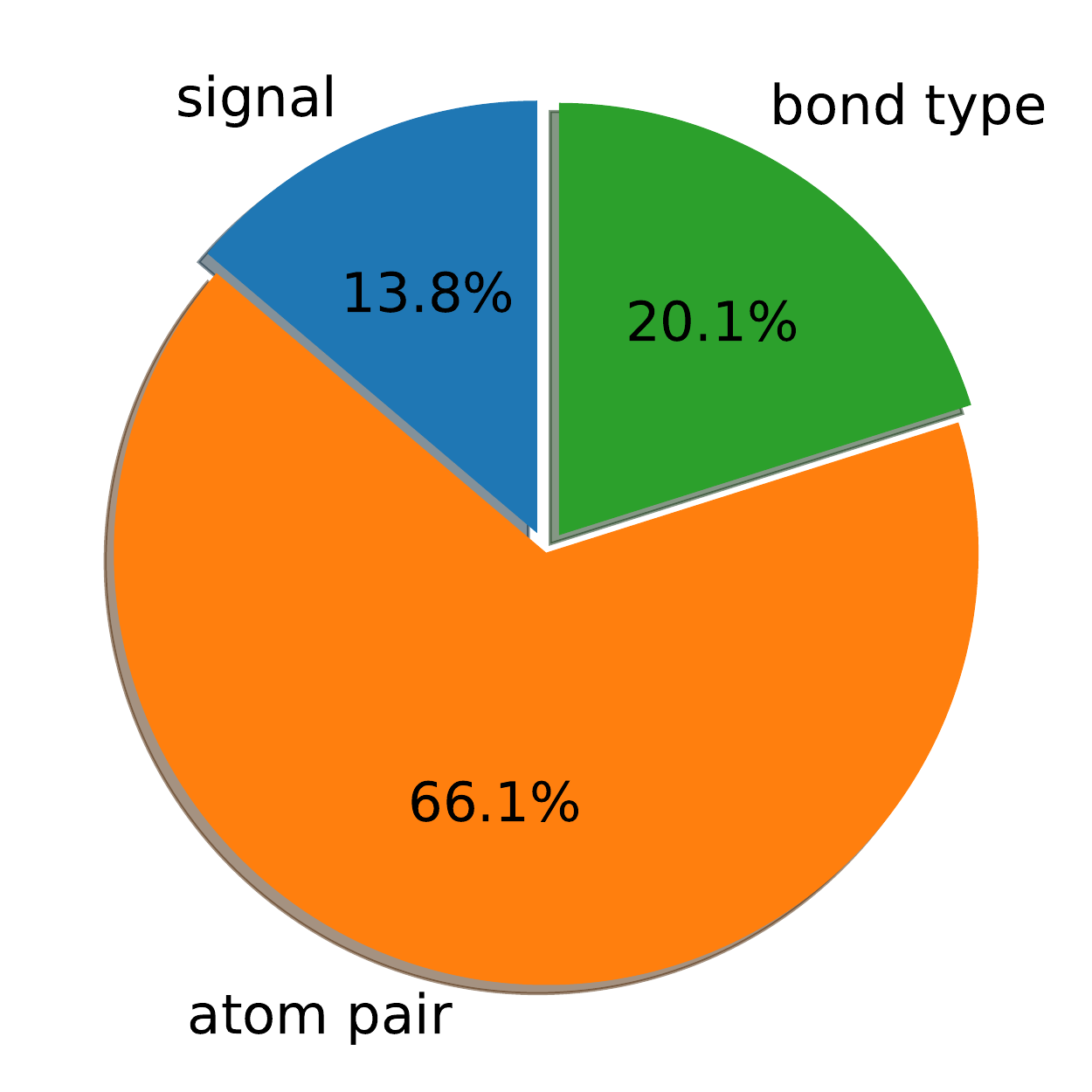}
\par\end{centering}
}\hspace*{0.15\textwidth}\subfloat[Proportion of the incorrect top-1 products that have similar structure
to the groundtruth products.\label{fig:error_symmetry}]{\includegraphics[width=0.4\textwidth]{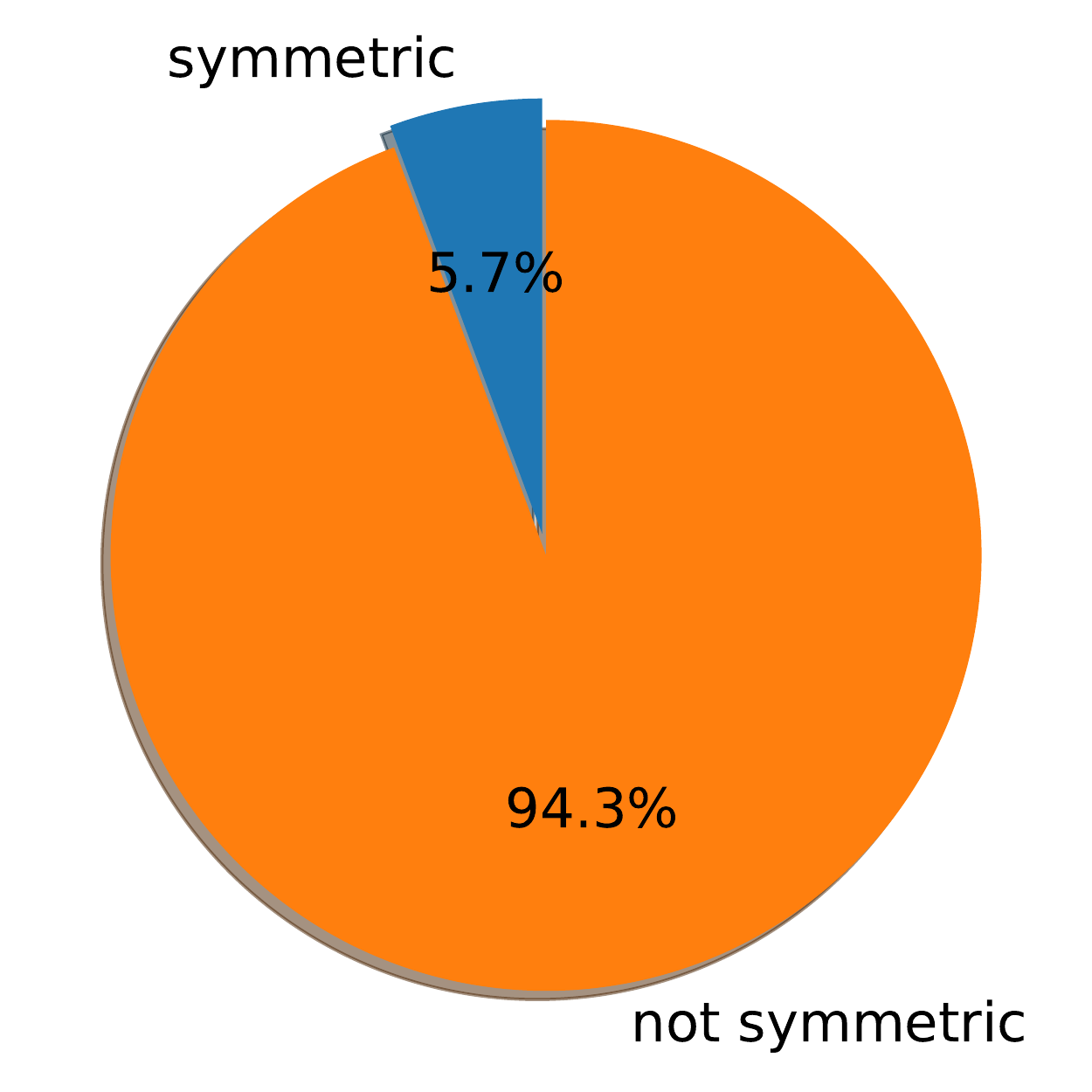}

}
\par\end{centering}
\caption{Errors grouped by the first incorrect sub-actions (a), and errors
caused by symmetric structures (b).}
\end{figure}

We define a sub-action causing error as the first sub-action that
our model makes a wrong decision. In Fig.~\ref{fig:error_subaction},
we plot the the proportion of errors with respect to the three kinds
of sub-actions. Clearly, atom pair prediction causes the most errors
(nearly two third). This makes sense because this sub-action is harder
than signal prediction and bond type prediction. Therefore, more effort
should be put on improving the prediction of atom pairs. 

\paragraph{Errors caused by symmetry}

\begin{figure}
\begin{centering}
\includegraphics[width=0.95\textwidth]{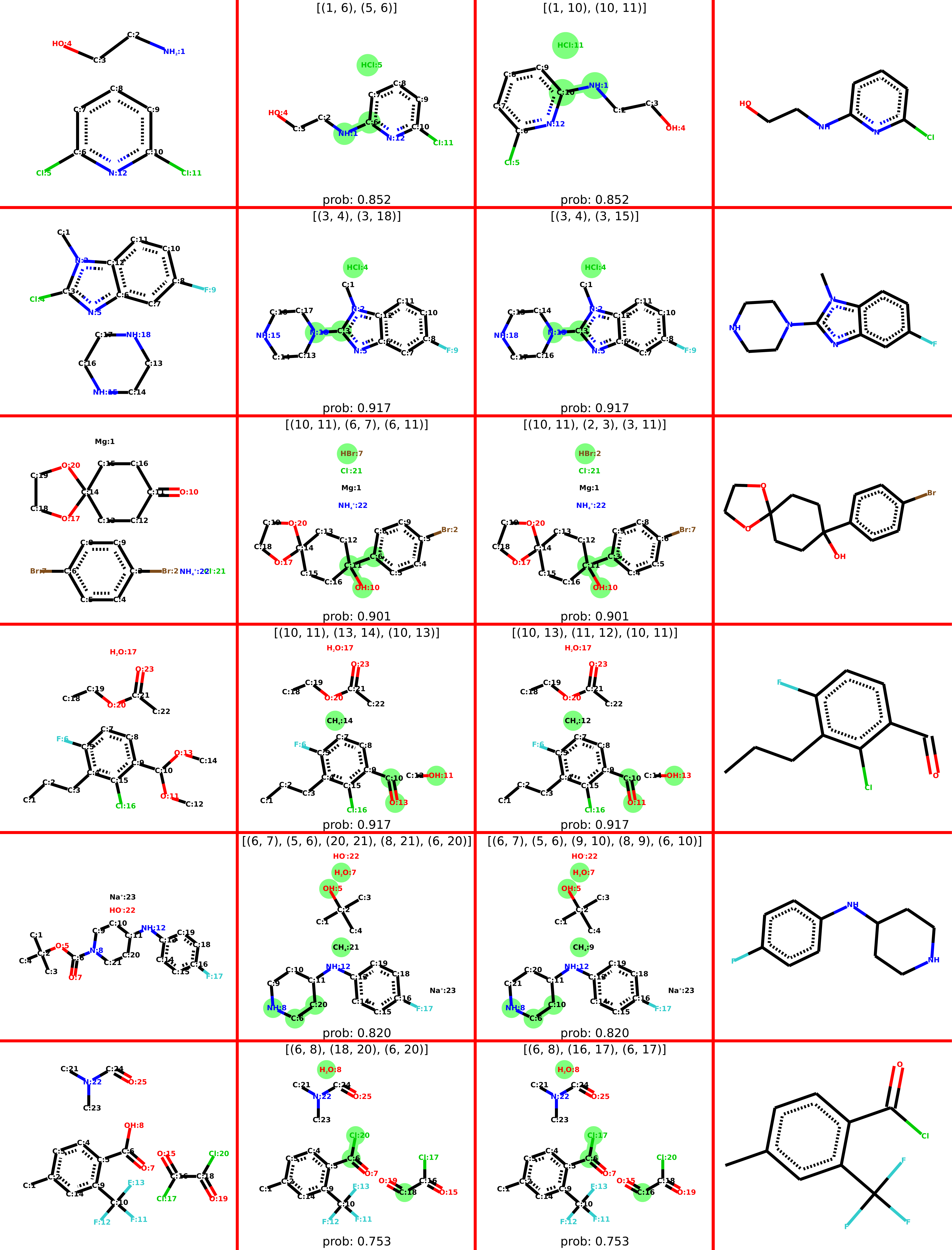}
\par\end{centering}
\caption{Visualization of some reactions that cause multiple products with
symmetric structures. Each row corresponds to a reaction. The columns,
from left to right, show: i) reactant and reagent molecules, ii) \textbf{incorrect}
top-1 product molecules, iii) \textbf{correct} top-2 product molecules,
and iv) major groundtruth product molecules. All atoms in the first
three columns are labeled with their atom map numbers. For the top-1
and top-2 products, we highlight the predicted reaction triples in
green and provide the probability of the predicted sequence at the
bottom.\label{fig:sym_prod_list}}
\end{figure}

There exists cases in which different sequences of bond changes can
result in the same products due to symmetric graph structures. Errors
caused by symmetry account for 5.7\% of the top-1 errors on the \emph{USPTO-15k}
dataset as shown in Fig.~\ref{fig:error_symmetry}. For better understanding,
we provide a short list of wrong reaction triple predictions caused
by symmetry in Fig.~\ref{fig:sym_prod_list}. In this list, the top-1
products (along the second column) are incorrect while the top-2 products
(along the third column) are correct though both have the same probability. 

\end{document}